\documentclass[sigconf]{acmart}
\acmSubmissionID{3332}

\AtBeginDocument{%
  \providecommand\BibTeX{{%
    \normalfont B\kern-0.5em{\scshape i\kern-0.25em b}\kern-0.8em\TeX}}}

\copyrightyear{2022}
\acmYear{2022}
\setcopyright{rightsretained}
\acmConference[MIG '22]{ACM SIGGRAPH Conference on Motion, Interaction and Games}{November 3--5, 2022}{Guanajuato, Mexico}
\acmBooktitle{ACM SIGGRAPH Conference on Motion, Interaction and Games (MIG '22), November 3--5, 2022, Guanajuato, Mexico}
\acmDOI{10.1145/3561975.3562943}
\acmISBN{978-1-4503-9888-6/22/11}

\usepackage{enumitem}
\usepackage{subcaption}
\usepackage[ruled,linesnumbered]{algorithm2e}
\usepackage{float}
\floatstyle{plaintop}
\restylefloat{table}
 \usepackage[nameinlink]{cleveref}
\usepackage{pifont}

\creflabelformat{equation}{#2\textup{#1}#3}
\crefname{equation}{eq.}{eqs.}
\Crefname{equation}{Eq.}{Eqs.}
\crefname{table}{table}{tables}
\Crefname{table}{Table}{Tables}
\crefname{figure}{figure}{figures}
\Crefname{figure}{Figure}{Figures}
\Crefrangeformat{equation}{Eqs.~#3#1#4--#5#2#6}
\crefrangeformat{figure}{Figs.~#3#1#4--#5#2#6}
\crefrangeformat{theorem}{Thms.~#3#1#4--#5#2#6}

\newcommand{\bs}{\mathbf{s}}
\newcommand{\ba}{\mathbf{a}}
\newcommand{\bq}{\mathbf{q}}
\newcommand{\bz}{\mathbf{z}}
\newcommand{\cmark}{\text{\ding{51}}}
\newcommand{\xmark}{\text{\ding{55}}}

\begin{document}

\title{Learning High-Risk High-Precision Motion Control}
\titlenote{Project website: \url{https://namheegordonkim.github.io/scoot-mig2022/}}

\author{Nam Hee Kim}
\email{namhee.kim@aalto.fi}
\orcid{0000-0002-9172-624X}
\affiliation{%
  \institution{Aalto University}
  \streetaddress{Konemiehentie 2}
  \city{Espoo}
  \country{Finland}
  \postcode{00076}
}

\author{Markus Kirjonen}
\email{markus.kirjonen@aalto.fi}
\orcid{0000-0002-4382-2165}
\affiliation{%
  \institution{Aalto University}
  \streetaddress{Konemiehentie 2}
  \city{Espoo}
  \country{Finland}
  \postcode{00076}
}

\author{Perttu H\"am\"al\"ainen}
\email{perttu.hamalainen@aalto.fi}
\orcid{0000-0001-7764-3459}
\affiliation{%
  \institution{Aalto University}
  \streetaddress{Konemiehentie 2}
  \city{Espoo}
  \country{Finland}
  \postcode{00076}
}

\renewcommand{\shortauthors}{Kim, et al.}

\begin{abstract}
Deep reinforcement learning (DRL) algorithms for movement control are typically evaluated and benchmarked on sequential decision tasks where imprecise actions may be corrected with later actions, thus allowing high returns with noisy actions. In contrast, we focus on an under-researched class of \emph{high-risk, high-precision} motion control problems where actions carry irreversible outcomes, driving sharp peaks and ridges to plague the state-action reward landscape. Using computational pool as a representative example of such problems, we propose and evaluate State-Conditioned Shooting (SCOOT), a novel DRL algorithm that builds on advantage-weighted regression (AWR) with three key modifications: 1) Performing policy optimization only using elite samples, allowing the policy to better latch on to the rare high-reward action samples; 2) Utilizing a mixture-of-experts (MoE) policy, to allow switching between reward landscape modes depending on the state; 3) Adding a distance regularization term and a learning curriculum to encourage exploring diverse strategies before adapting to the most advantageous samples. We showcase our features' performance in learning physically-based billiard shots demonstrating high action precision and discovering multiple shot strategies for a given ball configuration.

\end{abstract}

\ccsdesc[500]{Computing methodologies~Machine learning algorithms}
\ccsdesc[500]{Computing methodologies~Physical simulation}
\keywords{
reinforcement learning, 
trajectory optimization,
motion control
}

\begin{teaserfigure}
  \centering
  \includegraphics[width=0.99\textwidth, trim=0.1cm 13.0cm 0.1cm 13.5cm,clip]{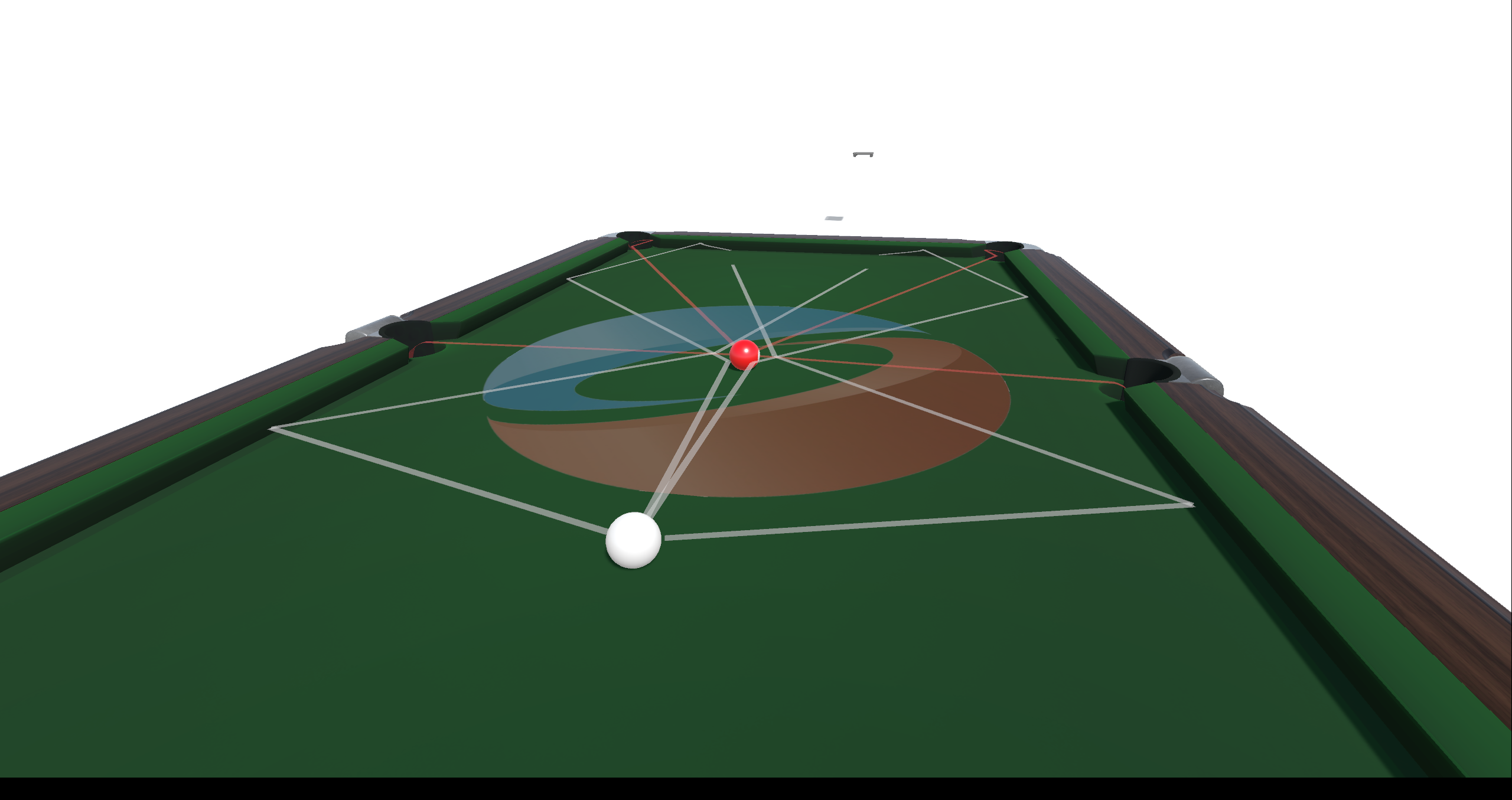}
  \caption{
Given its high-precision requirement and multimodality of solutions, learning shots for simulated billiards (a.k.a. computational pool) represents a long-standing challenge that has yet to be solved. A collection of neural network policies trained with our novel approach can produce multiple highly-skilled billiards shots for a given situation. The image shows four different shots sampled from our policies, given the same initial configuration.
  }
  \label{fig:teaser}
\end{teaserfigure}

\maketitle

\section{Introduction}


\begin{figure}
    \centering
    \begin{subfigure}{0.49\linewidth}
    \centering
    \includegraphics[width=0.7\textwidth,trim=0 0 0 17.5cm,clip]{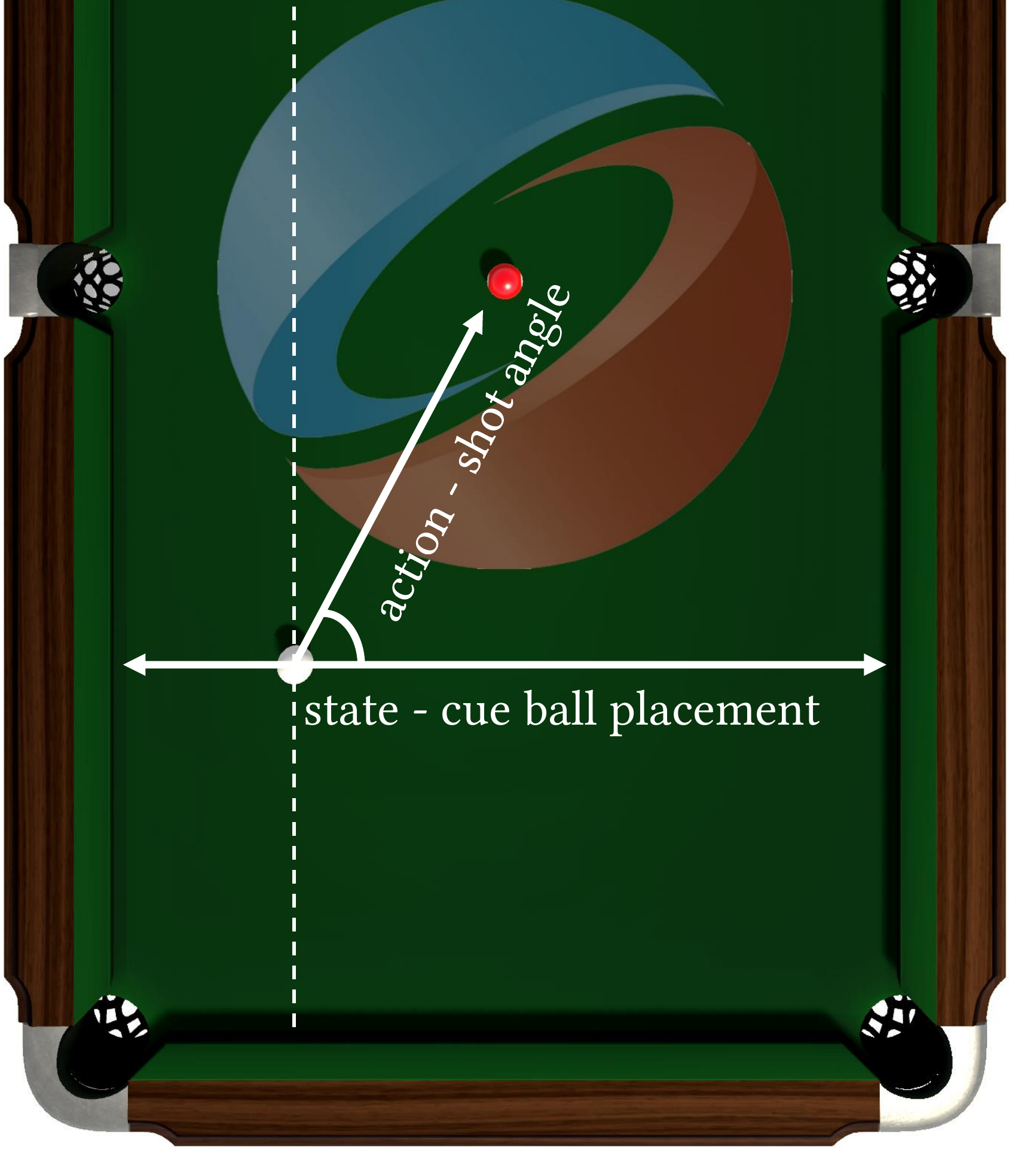}
    \end{subfigure}
    \begin{subfigure}{0.49\linewidth}
    \centering
    \includegraphics[width=1.0\textwidth,trim=5cm 0 5cm 0,clip]{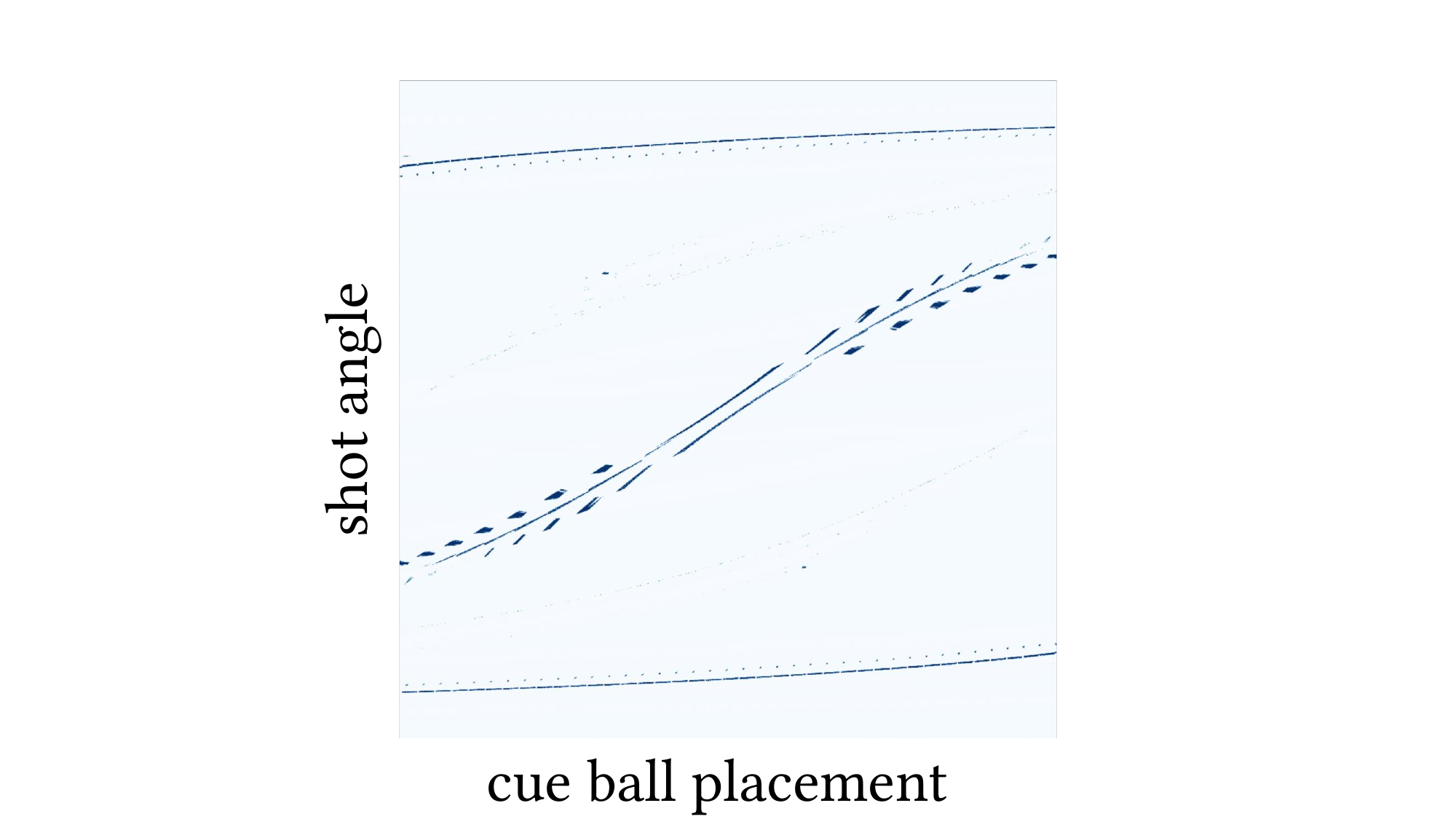}
    \end{subfigure}
    \caption{
    Left: a view of a physically-simulated billiards table with 1-dimensional state and action variables.
    Right: the state-action \emph{reward landscape} corresponding to the simplified billiards environment.
    Dark regions indicate high reward at the state-action coordinate.
    Best viewed in colour.
    }
    \label{fig:table_vis}
\end{figure}


\begin{figure*}
    \centering
    \begin{subfigure}[t]{0.25\textwidth}
    \captionsetup{justification=centering}
    \centering
    \includegraphics[width=\textwidth]{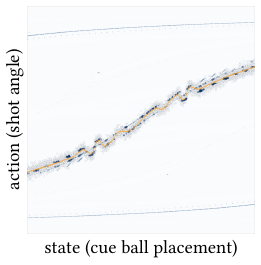}
    \caption{PPO}
    \label{fig:ppo_bad}
    \end{subfigure}\hfill
    \begin{subfigure}[t]{0.25\textwidth}
    \captionsetup{justification=centering}
    \centering
    \includegraphics[width=\textwidth]{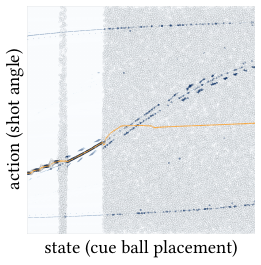}
    \caption{SAC}
    \label{fig:sac}
    \end{subfigure}\hfill
    \begin{subfigure}[t]{0.25\textwidth}
    \captionsetup{justification=centering}
    \centering
    \includegraphics[width=\textwidth]{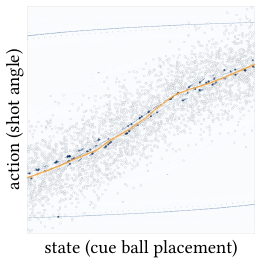}
    \caption{AWR\hspace{\linewidth}(fixed action var.)}
    \label{fig:awr_na_bad}
    \end{subfigure}\hfill
    \begin{subfigure}[t]{0.25\textwidth}
    \captionsetup{justification=centering}
    \centering
    \includegraphics[width=\textwidth]{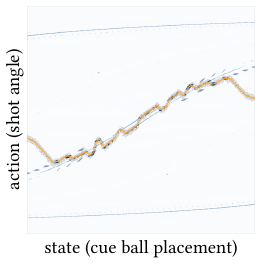}
    \caption{AWR\\(learned action var.)}
    \label{fig:awr_la_bad}
    \end{subfigure}\hfill
    \caption{Visualization of converged policies on the one-dimensional billiards reward landscape. Orange markers indicate state-conditioned action means, and coloured circles indicate state-action samples in the replay buffer, with hues indicating observed returns. Best viewed in colour.}
    \label{fig:prev_bad}
\end{figure*}

Deep reinforcement learning (DRL) is today's dominant paradigm for controlling physically simulated characters in computer animation and robotics \cite{bergamin2019drecon,ScaDiver,merel2020catch, peng2021amp}, superseding previous techniques such as direct space-time optimization \cite{witkin1988spacetime} and trajectory optimization techniques based on gradient information \cite{tassa2012synthesis,tassa2014control} or sampling \cite{al2012trajectory, liu2012terrain, hamalainen2014online, hamalainen2015online}.
Although DRL can produce highly unrealistic movements in common benchmarks such as OpenAI Gym \cite{brockman2016openai} and DeepMind Control Suite \cite{tassa2018deepmind}, this can be explained by the simplistic benchmark simulation models and reward functions. Guiding DRL with suitable reference movement data can help synthesize natural and controllable movements \cite{bergamin2019drecon,ScaDiver,merel2020catch, peng2019advantage}.


Thus far, the considerable body of work on DRL in continuous action spaces studies a fairly limited set of test problems. In particular, locomotion, balancing, and other common sequential control tasks allow the agent to produce occasional noisy or erroneous actions, which later actions can correct by compensating for the noise/error. The same does not apply to many other tasks like shooting an arrow or hitting a sculpture with a chisel. In such scenarios, the agent has little to no control over the outcome once the action is executed, making the task less error-tolerant and hence less forgiving. Consequentially, the learned policy must be much more precise. Motivated by this, we pose and tackle the following question: \emph{Can DRL algorithms solve motion control problems that are unforgiving towards imprecise actions?}

We examine the problem of billiard shot learning as a representative problem, capturing crucial properties of the class of problems above,  which we dub \emph{high-risk and high-precision} motion control problems. Successful shots in billiards require highly precise (low-entropy) actions, as even one or two degrees of shot angle difference can make a huge difference in the resulting trajectory. Also, failed shots cause drastic and irreversible state changes, which an opponent can exploit. Billiards is also a convenient testbed from a computational and didactic perspective---in its simplest form, it can be studied with only 1 or 2 state and action dimensions, allowing highly informative visualizations of the reward landscapes and algorithm behaviour.

In this paper, we first identify opportunities for algorithmic improvement by examining the performance of multiple off-the-shelf DRL algorithms. We then develop a set of algorithm modifications that allows for sufficient adaptation to the highly sparse, sharp, disjointed, and multimodal reward landscapes of billiards, thereby achieving highly precise control of the cue ball. As a result, we produce a neural network policy that can compute successful billiard shots in a single forward pass, which is a significant step-up from existing solutions based on iterative optimization techniques such as covariance matrix adaptation evolution strategy (CMA-ES, \citealt{hansen2003reducing}).

Our contributions can be summarized as follows: 
\begin{itemize}[leftmargin=*]
    \item We test and visualize several off-the-shelf DRL algorithms (PPO \cite{schulman2017proximal}, SAC \cite{haarnoja2018soft}, and AWR \cite{peng2019advantage}) and show that they fail even in the most simple billiards variants. We analyze the visuals and highlight opportunities for improvement.
    \item Novel algorithm: we propose and evaluate a novel DRL approach named SCOOT (\textbf{S}tate-\textbf{C}onditioned Sh\textbf{oot}ing) that combines features from existing algorithms and modifications motivated by the characteristics of billiards.
    \item We demonstrate the capacity of SCOOT to both produce highly precise shots and suggest multiple alternative strategies for a given ball configuration. 
\end{itemize}

\section{Related Work}

\paragraph{Sampling-based Optimization for Control}

Sampling-based optimization is a well-suited tool for solving high-risk, high-precision motion control problems. Combined with physical simulation, sampling-based optimization has proven successful in performing physics-based motion control as a form of shooting method. Basic tools for optimizing action sequences range from the cross-entropy method (CEM, \citealt{de2005tutorial}) and covariance matrix adaptation evolution strategy (CMA-ES, \citealt{hansen2003reducing}) to more recent methods such as iCEM \cite{pinneri2020sample}. Trajectory optimization with a long simulation horizon has been shown to induce requirements for high-precision \cite{hamalainen2020visualizing}, where a high ratio between simulation and control frequencies warrants that a controller must construct precise action sequences. Sampling-based trajectory optimization has successfully performed several trajectory optimization tasks with careful reward engineering \cite{al2012trajectory, naderi2017discovering}. More recently, CMA-ES was used to perform physics-based keyframe tracking as a space-time optimization solver \cite{witkin1988spacetime} in conjunction with compact action parameterization \cite{kim2021flexible}. The main problem hindering the application of sampling-based optimization in practice is computing cost and time delay in obtaining a solution. Hence, the focus of computer animation and movement control research has shifted to learning neural control policies that compute motion control parameters in a single forward pass, often requiring mere milliseconds of GPU time. DRL is closely related to sampling-based optimization: one may view learning a neural policy as solving multiple variants of a movement optimization problem in parallel, each variant specified by the policy network's input \cite{hamalainen2020ppo}. Typically, this means that optimized variables such as the mean and variance of CMA-ES search distribution become functions of the policy network input, parameterized by the network weights and biases.

\paragraph{Deep Reinforcement Learning}
\citet{kwiatkowski2022survey} presents a comprehensive survey of reinforcement learning methods for character animation. The first steps for benchmarking physics-based control tasks and environments often involve popular off-the-shelf algorithms such as Proximal Policy Optimization (PPO, \citealt{schulman2017proximal}) and Soft Actor-Critic (SAC, \citealt{haarnoja2018soft}). Modifications for these off-the-shelf algorithms, such as Advantage-Weighted Regression (AWR, \citealt{peng2019advantage}), have been proposed over the years, showing potential performance improvements. More recently, mixture-of-experts (MoE) methods \cite{peng2019mcp,won2020scalable} have gained traction due to their capabilities to model diverse motions in the presence of multiple objectives or multi-task motion data. Our work is closely related to AWR in viewing DRL as a weighted regression task while employing MoE for performance improvement. Complementing previous MoE work, we contribute novel visualizations of the reward landscape that help explain why and when MoE policies can be useful.

\nocite{bergamin2019drecon}  
\nocite{ScaDiver}  

\paragraph{Computational Pool}
Computational pool is a great testbed for optimization, noted for the difficulties inherent to planning in continuous domain \cite{greenspan2005uofa, landry2011billiards}.
Previous attempts at solving billiards include Monte-Carlo tree search approaches \cite{smith2007pickpocket, chen2019macro} and variants of sampling-based trajectory optimization \cite{landry2007ai, archibald2010computational}. \citet{fragkiadaki2015learning} successfully applied CMA-ES in conjunction with a visual predictive model of the pool table with multiple balls. However, learning to produce viable shots in real-time has remained an open challenge. \citet{hu2019difftaichi} presents a differentiable billiards simulator, backpropagating through which can optimize a shot. However, their simplified setup excludes table borders, and the gradient-based shot optimization can take multiple iterations depending on initialization. To our best knowledge, \citet{el2019recommender} is the only publicly available academic work that solves the problem using a DRL algorithm. However, this work treats billiards as a sequential decision (i.e. multiple shots are allowed in sequence) rather than a high-risk high-precision problem.

\section{Notation and Preliminaries}

As usual with DRL, we assume a Markov Decision Process (MDP) where at timestep $t$, the agent observes a state $\bs_t$ and takes an action $\ba_t$, which results in a new state $\bs_{t+1}$ and scalar reward $r_t=r(\bs_t,\ba_t)$. A policy network $\pi_\theta$ defines a state-conditioned distribution for sampling actions as $\ba \sim \pi_\theta (\ba | \bs)$, where $\theta$ denotes policy parameters. In our case, $\bs$ defines the initial ball configuration for a billiards shot, and $\ba$ defines the cue ball's launch velocity.%

Other key concepts include the return $R(\bs,\ba)=\sum_t \gamma^t r_{t}$ of a state-action sequence starting at state $\bs=\bs_0$ and action $\ba=\ba_0$, where $\gamma \in [0,1)$ is a discounting parameter. State value or expected return for a state is denoted by $V(\bs)=\mathbb{E}_\ba [R(\bs,\ba)]$. The objective is to maximize the expected return over some state distribution, in our case the positions of billiards balls sampled randomly. In this paper, we only focus on learning single optimal shots instead of shot sequences, setting $\gamma=0$. This makes the returns equal to instantaneous rewards, $R(\bs,\ba)=r(\bs,\ba)$, which can simplify DRL algorithm implementation.%

\section{Billiards Simulator} \label{sec:simulator}

For our experiments, we implement a high-fidelity billiards environment in Unity with NVIDIA PhysX engine in the style of OpenAI's Gym interface. The placement of the cue ball is parameterized into the state vector, and the agent is to output action parameters which apply an impulse to the cue ball at the beginning of the simulation. The rest of the trajectory follows from contacts and collisions accrued as a consequence. For exploring the performances of off-the-shelf algorithms, we simplify state and action variations to just 1 dimension each. As shown in \Cref{fig:table_vis}, the output of the action is the shot angle, where the striking force is held constant. We make the problem a single decision rather than a sequential one and simulate for a long horizon (200 timesteps with a single action). This reflects how in real billiards, new actions are only taken after the balls have stopped moving after the previous action; thus, each action can have extensive and irreversible effects on the game state.

\paragraph{Reward Design} The reward function reflects the characteristics of the whole shot trajectory and use simple bonus and penalty terms bounded in $[0, 1]$, which simplifies balancing their weights. Specifically, our reward function is as follows:
\begin{align}
    r(\bs, \ba) &= \nonumber
    \mathcal{I}\left( \text{target ball pocketed} \right) \nonumber \\
    &+ 0.01 \cdot \exp \left( -\min_{p, t} ||\bq^\text{target}_t - \bq^p||^2 \right) \nonumber \\
    &+ 0.01 \cdot \left(1 - \exp \left( -\min_{p, t} ||\bq^{\text{cue}}_t - \bq^p||^2 \right)\right) \label{eq:reward} 
\end{align}%
where $\mathcal{I}(\cdot)$ is the indicator function, $\bq^{\text{cue}}_t$ and $\bq^{\text{target}}_t$ are the $xy$-positions of the cue ball and the target ball at the physics timestep $t$ of the simulated shot trajectory, and $\bq^p$ is the $xy$-position of the pocket $p$. In case the cue ball is pocketed, the trajectory registers as a failure, and the reward is set to 0. The exponential terms are reward shaping terms that guide the target ball to move close to a pocket while penalizing for the cue being too close to any pocket. The weights of the shaping terms are set to be low in relation to the binary pocketing reward so that the maximum reward cannot be obtained without an actually successful shot. Without the shaping terms, the reward landscape would have large regions with zero gradient.

\paragraph{Visualizing the Optimization Landscape} In the case of 1-dimensional states and actions, we can directly visualize the policy optimization problem using $s,a$ as the axes, $r(s,a)$ as the objective function. Actions sampled from the policy can also be plotted on top of the objective function landscape to visualize DRL algorithm progress and convergence. Examples of this are provided in \Cref{fig:table_vis} and \Cref{fig:prev_bad}. One can see that the high-precision action requirement manifests as the landscape having narrow peaks and ridges. A good policy is one that samples actions only at the peaks/ridges, for any given state. 

The fact that one can aim at different pockets and utilize bounces manifests as multimodality---most states have more than a single high-reward action. Between states, the high-reward actions change somewhat smoothly, but there are also discontinuities resulting from some shots pocketing the cue ball together with the target ball. 

\section{Problems with Off-the-Shelf DRL}
Before introducing our SCOOT algorithm, let us first establish that the problem of learning billiards shots is indeed hard and not solved by common off-the-shelf DRL algorithms. We exclude model-based methods from consideration due to the high sensitivity of state transitions arising from billiards' contact- and collision-rich nature. As such, leveraging learned dynamics models becomes infeasible: prediction errors for state transitions will necessarily result in imprecise actions.
Here, we examine the performance of three popular model-free, actor-critic based DRL methods, namely, Proximal Policy Optimization (PPO, \citealt{schulman2017proximal}), Soft Actor-Critic (SAC, \citealt{haarnoja2018soft}), and Advantage-Weighted Regression (AWR, \citealt{peng2019advantage}). For PPO and SAC, we utilize the popular Stable Baselines 3 implementations \cite{raffin2019stable}. For AWR, we use the official paper codebase. 
As visualized in \Cref{fig:prev_bad}, even in the simplest case (1-dimensional state and action), these methods, without further modification, fail to achieve a viable solution across the state space:
\begin{itemize}[leftmargin=*]
    \item PPO (\Cref{fig:ppo_bad}) locks on to the discontinuities of the reward landscape, resulting in the learned policy fluctuating between landscape peaks. There are multiple states where the learned policy produces unsuccessful shots.
    \item SAC (\Cref{fig:sac}) fails to lower its state-conditioned action entropy adequately, possibly due to the entropy regularization mechanic that biases the algorithm more towards exploration than exploitation, which is not a good fit for high-risk, high-precision problems. 
    \item AWR without learned action variance (the default, \Cref{fig:awr_na_bad}) is only able to model the rough shape of the peaks present in the reward landscape. With learned action variance (\Cref{fig:awr_la_bad}), AWR adapts more appropriately to the peaks, but the final policy fluctuates between peaks similar to PPO. However, here is less action noise, possibly because of the ability to use a large buffer of off-policy experience. Because of the sharpness of the reward peaks, high-reward actions are sampled only rarely, and using off-policy data from previous iterations helps AWR to not "forget" the good actions after each iteration, allowing the algorithm to model the reward landscape peaks more accurately.
\end{itemize}

Throughout the rest of the paper, we use AWR with learned action variance as the baseline and starting point when developing and evaluating our SCOOT algorithm.%

\section{State-Conditioned Shooting}


\begin{table*}[]
\begin{tabular}{lllllll}
\hline
\textbf{Algorithm} & \textbf{On/Off Policy} & \textbf{Action Distr.} & \textbf{Elite Selection?} & \textbf{Action Var.} & \textbf{Curriculum?} \\ \hline
PPO \cite{schulman2017proximal}                & On              & Gaussian             & {\xmark} & Learned global & {\xmark}                    \\
SAC \cite{haarnoja2018soft}                & Off             & Gaussian              & {\xmark}     & State-conditioned & {\xmark}                     \\
AWR \cite{peng2019advantage}                & Off             & Gaussian              & {\xmark}  & Fixed (default) or learned global & {\xmark}                   \\
\textbf{SCOOT (ours)}       & Off             & Gaussian Mixture   & {\cmark}   & Learned global & {\cmark}
\end{tabular}
\caption{Summary of differences between SCOOT (ours) and commonly-used off-the-shelf model-free DRL algorithms.}
\label{table:comparisons}
\end{table*}

Our proposed SCOOT algorithm builds on AWR, with modifications proposed and evaluated in the following sections. \Cref{fig:incremental,fig:learning_curve} summarize the results of each proposed modification, showing both the converged policies and the learning curves, i.e., the learned policy's mean return as a function of the number of training shots.

\begin{algorithm*}
\caption{State-Conditioned Shooting (SCOOT)}
\label{alg:scoot}
\SetKwProg{scoot}{Function ScootStep}{}{end}
\SetKwFor{RepTimes}{repeat}{times}{end}
$\mathcal{D} \leftarrow$ Experience replay buffer, initially empty FIFO queue\\
$\theta, \phi \leftarrow$ Parameters of policy ($\pi_\theta$) and value ($V_\phi$) networks \\
\For{\text{iteration} $k=1, \cdots, k_\text{max}$}{
add experience $\left\{\left< \bs, \ba, R(\bs,\ba) \right>_{i, \cdots, n}\right\}$ sampled via $\pi_\theta$ to $\mathcal{D}$ \\
$\phi \leftarrow \arg\min_\phi \mathbb{E}_{\left<\bs, \ba\right> \sim \mathcal{D}}\left[ ( R(\bs,\ba) - V_\phi(\bs) )^2 \right]$ \tcp*{Train $V_\phi$ (\Cref{eq:critic_loss})} 
$\mathcal{D}^* \leftarrow$ SelectElites$\left(\mathcal{D}, \phi\right)$ \label{algline:elite} \tcp*{(\Cref{eq:elite})}
$\theta \leftarrow \arg\max_\theta \mathbb{E}_{\left<\bs, \ba\right> \sim \mathcal{D}^*}\Bigl[ \underbrace{\vphantom{\lambda_{\text{dist}}}A(\bs, \ba) \log\pi_\theta\left( \ba|\bs \right)}_{\text{Advantage-weighted regression}} - \underbrace{\lambda_{\text{dist}} \cdot \mathcal{L}_\text{dist}}_{\text{Distance regularization}}\Bigr]$ \label{algline:policy} \tcp*{Train $\pi_\theta$ (\Cref{eq:loglik,eq:moe_prob,eq:dist_loss})}
}
\end{algorithm*}

\begin{figure}
    \centering
    \includegraphics[width=0.99\columnwidth,trim={60 150 125 150},clip]{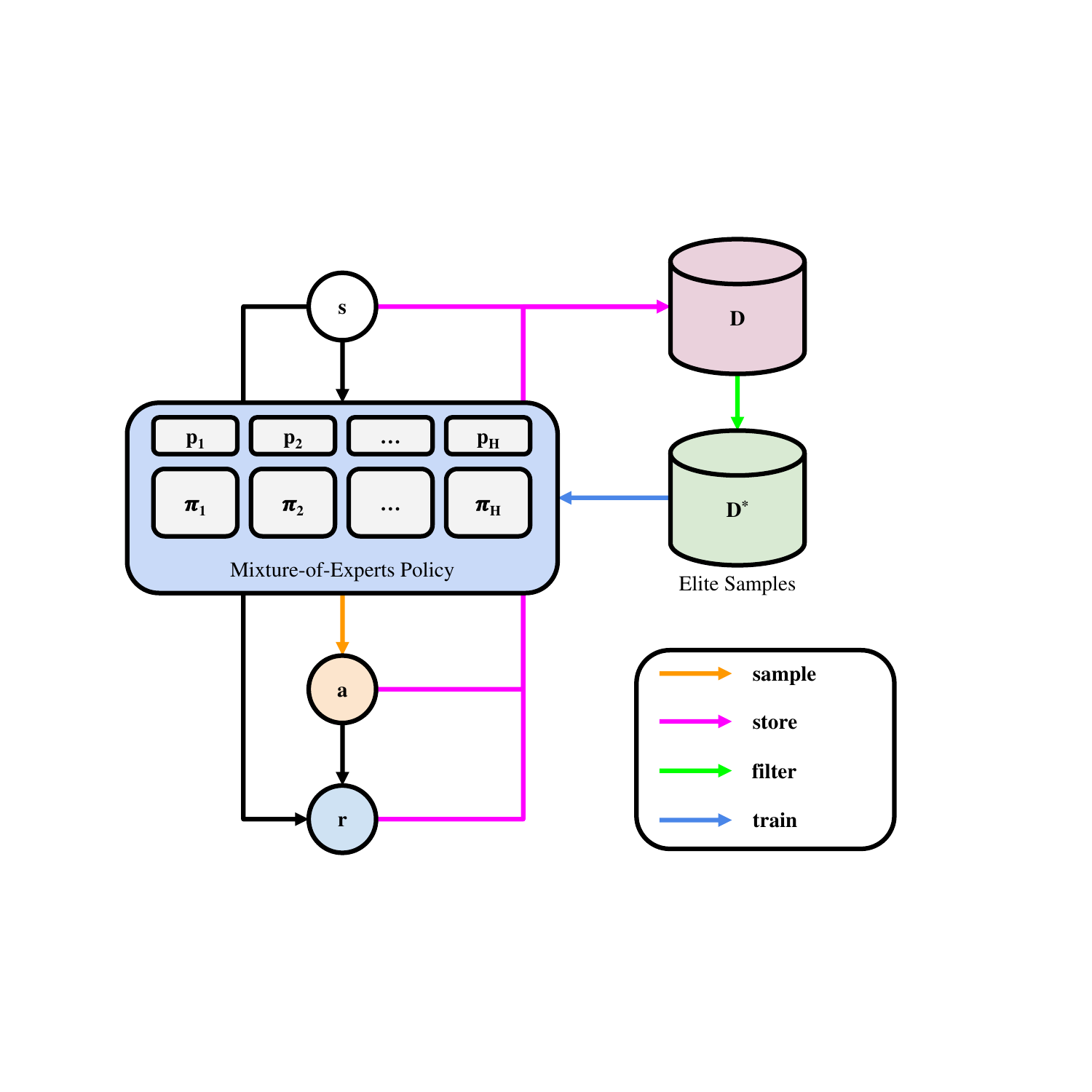}
    \caption{Overview of policy optimization iterations with a mixture-of-experts policy and elite samples from off-policy experience replay buffer.}
    \label{fig:overview}
\end{figure}

\paragraph{Algorithm overview} The key operating principle of both SCOOT and AWR is the same: \emph{At each iteration, new actions are sampled, simulated, and added to a FIFO experience buffer, and the policy distribution is fitted to the actions that yield high returns}.
This follows the trend of casting RL as a (self-)supervised learning problem \cite{eysenbach_kumar_gupta_2020,ghosh2019learning,kumar2019reward}. Conceptually, this is also similar to black-box optimization methods like CMA-ES and CEM, which implement a similar iteration of sampling candidate optimization solutions from a search distribution and fitting the distribution to the best candidates. However, the search distribution of CMA-ES and CEM is not conditioned on state, and they only use the current iteration's sampled experience in the distribution updates. In contrast, AWR's and SCOOT's policy networks define state-conditioned search distributions for actions, and a multi-iteration experience replay buffer (a FIFO queue) avoids forgetting old actions that yielded high returns, which is a good fit for our case where sampling good actions is rare. This comes at the cost of possibly slower convergence, as the policy can get "stuck" to old actions that are good but suboptimal; the AWR paper shows how fitting the policy to old experiences corresponds to optimizing a bound instead of the true objective. 

More specifically, the policy is fitted to the best actions by updating $\theta$ to maximize the weighted log-likelihood:
\begin{equation}
    \theta \leftarrow \arg\max_\theta \mathbb{E}_{\left<\bs,\ba\right> \sim \mathcal D  }\left[W(\bs, \ba) \log\pi_\theta\left( \ba|\bs \right)\right], \label{eq:loglik}
\end{equation}
where $\mathcal{D}$ denotes the experience replay buffer, and the weights $W(\bs,\ba)$ denote how good a sampled action $\ba$ is in state $\bs$. AWR uses $W(\bs,\ba)=\exp(\frac{1}{\beta} \cdot  A(\bs,\ba))$, where $\beta$ is a hyperparameter, and $A(\bs,\ba)$ is an advantage estimate $A(\bs,\ba)=R(\bs,\ba)-V(\bs)$. Intuitively, the advantage is positive if action $\ba$ yields a higher return than expected in state $\bs$. The exponentiation maps the advantages to non-negative sample weights. For a practical implementation, $V(\bs)$ is approximated by a value network $V_\phi(\bs)$ trained with the observed returns in a supervised manner, minimizing the L2-loss:
\begin{equation}
    \phi \leftarrow \arg\min_\phi \mathbb{E}_{\left<\bs,\ba\right> \sim \mathcal D  }\left[\left(R(\bs,\ba)-V_\phi(\bs)\right)^2\right].
\label{eq:critic_loss}
\end{equation}
The SCOOT algorithm is summarized in \Cref{fig:overview} and \Cref{alg:scoot}. This follows the AWR algorithm definition with the following modifications:
\begin{itemize}[leftmargin=*]
    \item We add the selection of elite samples (Line \ref{algline:elite}, \Cref{sec:elite})
    \item Enabled by the above, we also use advantages as weights without exponentiation, $W(\bs,\ba)=A(\bs,\ba)$ (Line \ref{algline:policy}, \Cref{sec:elite})
    \item We use a state-conditioned Gaussian mixture instead of a single Gaussian as the action distribution (\Cref{sec:moe})
    \item We add distance regularization to the policy update (Line \ref{algline:policy}, \Cref{sec:moe})
\end{itemize}
Additionally, we implement a 3-stage training curriculum to prevent premature convergence (\Cref{sec:curriculum}). \Cref{table:comparisons} provides a summary of the difference between SCOOT and other common model-free DRL algorithms. 

\subsection{Elite Samples} \label{sec:elite}
One key consequence of our reward landscape's sparsity and sharpness is the overwhelmingly high ratio between low-return and high-return action samples, especially at initialization. In fitting the policy, one would like to only focus on the high-return actions, disregarding the low-return ones. Such fitting to only \emph{elite samples}, e.g., top 50\%, is also central to black-box optimization algorithms such as CMA-ES and CEM. 

More formally, we select elite samples from the experience replay buffer, using the value network's state value estimate $V_\phi(\bs)$:
\begin{align}
    \mathcal{D}^* &= \left\{ \left<\bs, \ba, R(\bs,\ba)\right> \ | \ \left<\bs, \ba, R(\bs,\ba)\right> \in \mathcal{D}, R(\bs,\ba) > V_\phi(\bs)  \right\} \label{eq:elite}
\end{align}
A consequence of the above is that the advantage estimates $A(\bs,\ba)=R(\bs,\ba) - V_\phi(\bs)$ of the elite samples are always non-negative. Thus, the AWR-style exponentiation is no longer necessary to convert advantages to non-negative sample weights. 

\begin{figure*}
    \centering
    \begin{subfigure}[t]{0.18\textwidth}
    \centering
    \captionsetup{justification=centering}
    \includegraphics[width=1.0\textwidth]{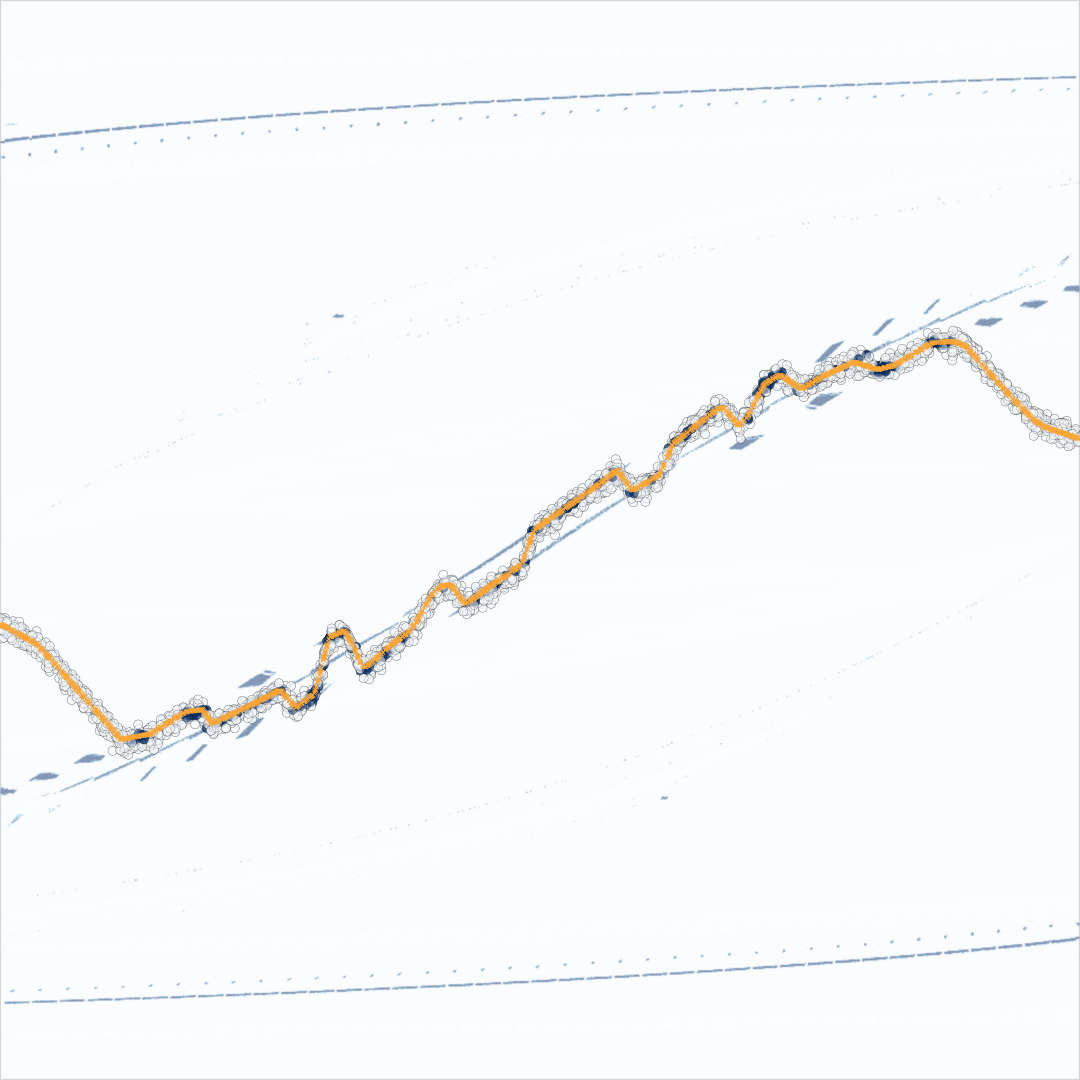}  
    \caption{Baseline AWR}
    \end{subfigure}\hfill
    \begin{subfigure}[t]{0.18\textwidth}
    \centering
    \captionsetup{justification=centering}
    \includegraphics[width=1.0\textwidth]{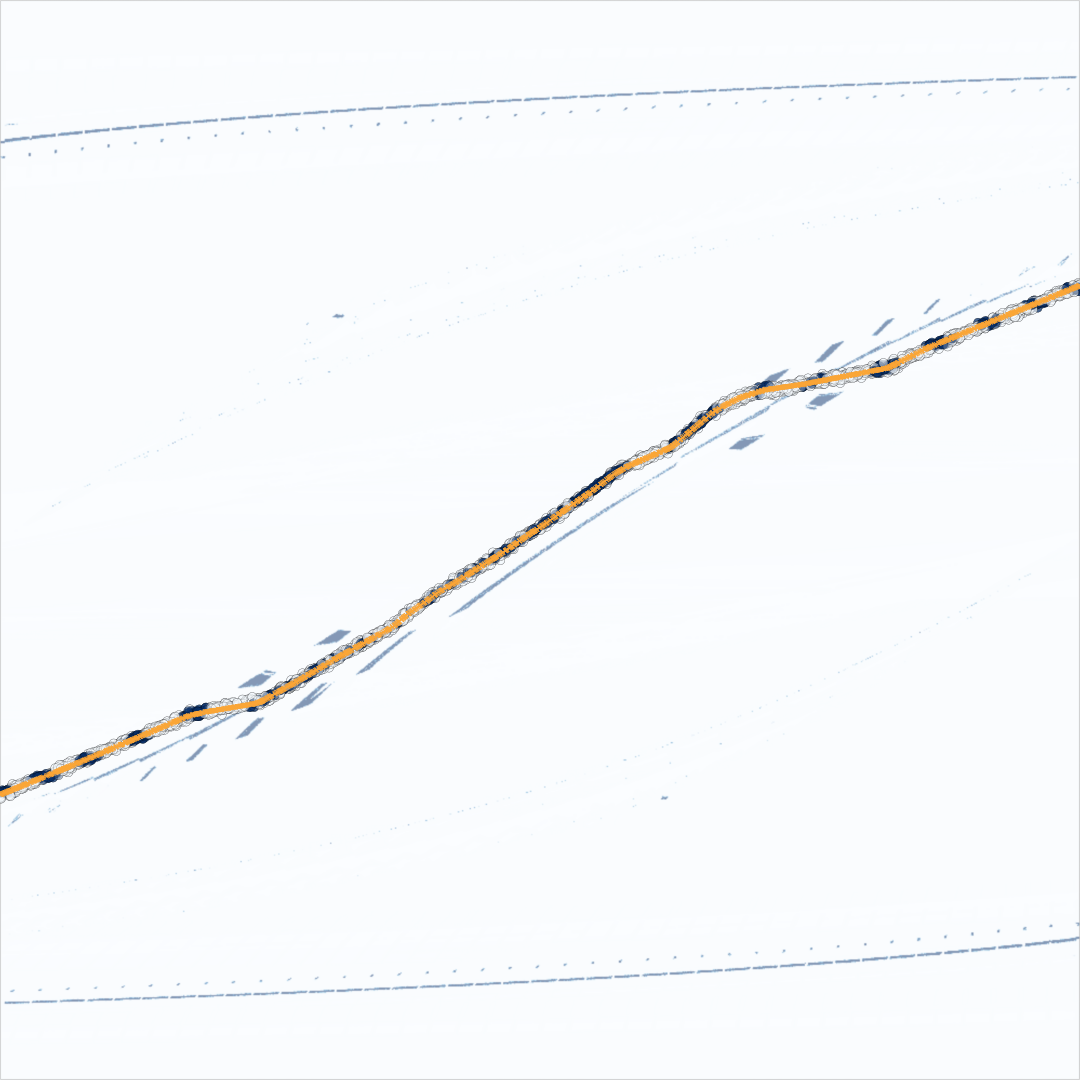}  
    \caption{AWR\\+~elite samples}
    \end{subfigure}\hfill
    \begin{subfigure}[t]{0.18\textwidth}
    \centering
    \captionsetup{justification=centering}
    \includegraphics[width=1.0\textwidth]{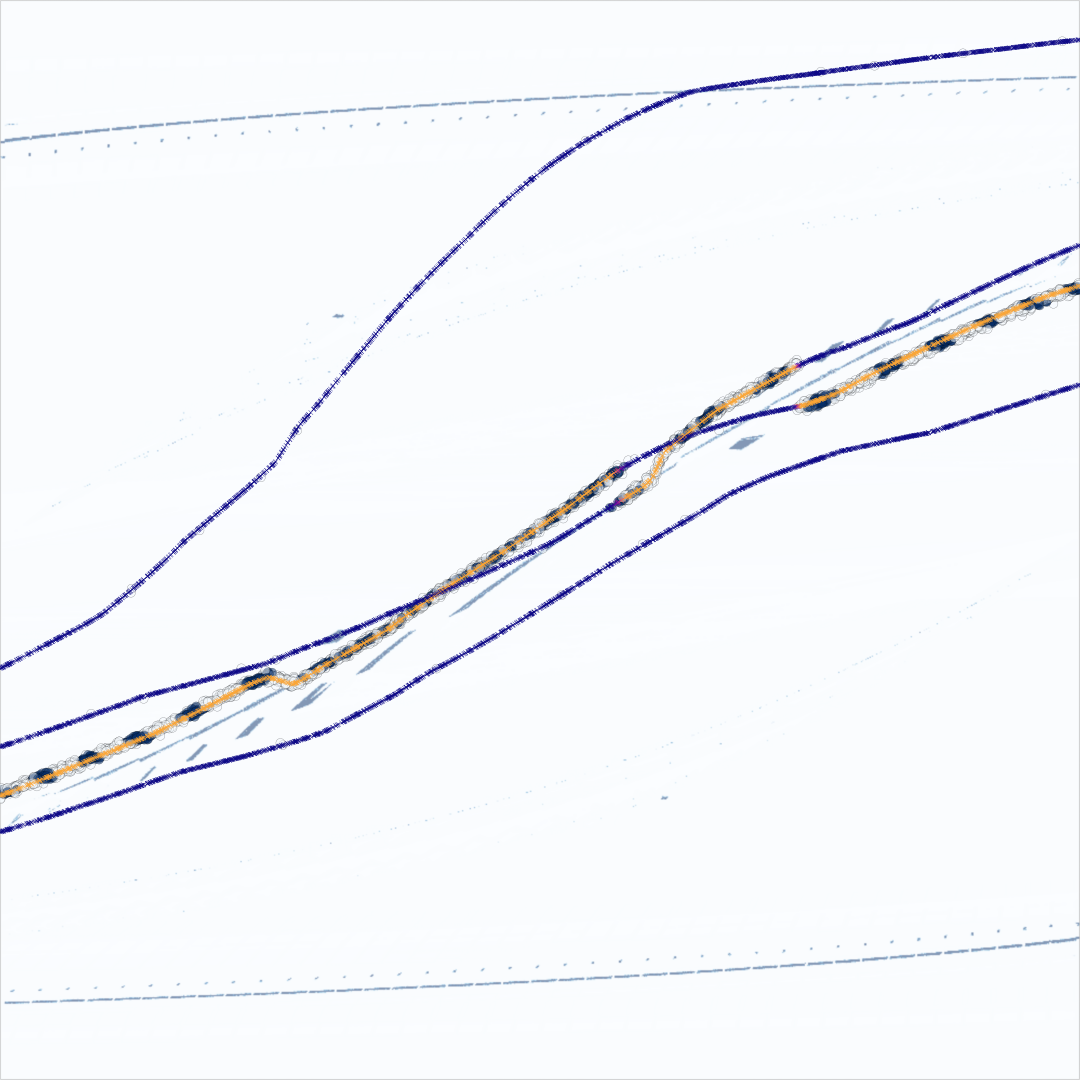}  
    \caption{Na\"ive MoE}\label{fig:naivemoe}
    \end{subfigure}\hfill
    \begin{subfigure}[t]{0.18\textwidth}
    \centering
    \captionsetup{justification=centering}
    \includegraphics[width=1.0\textwidth]{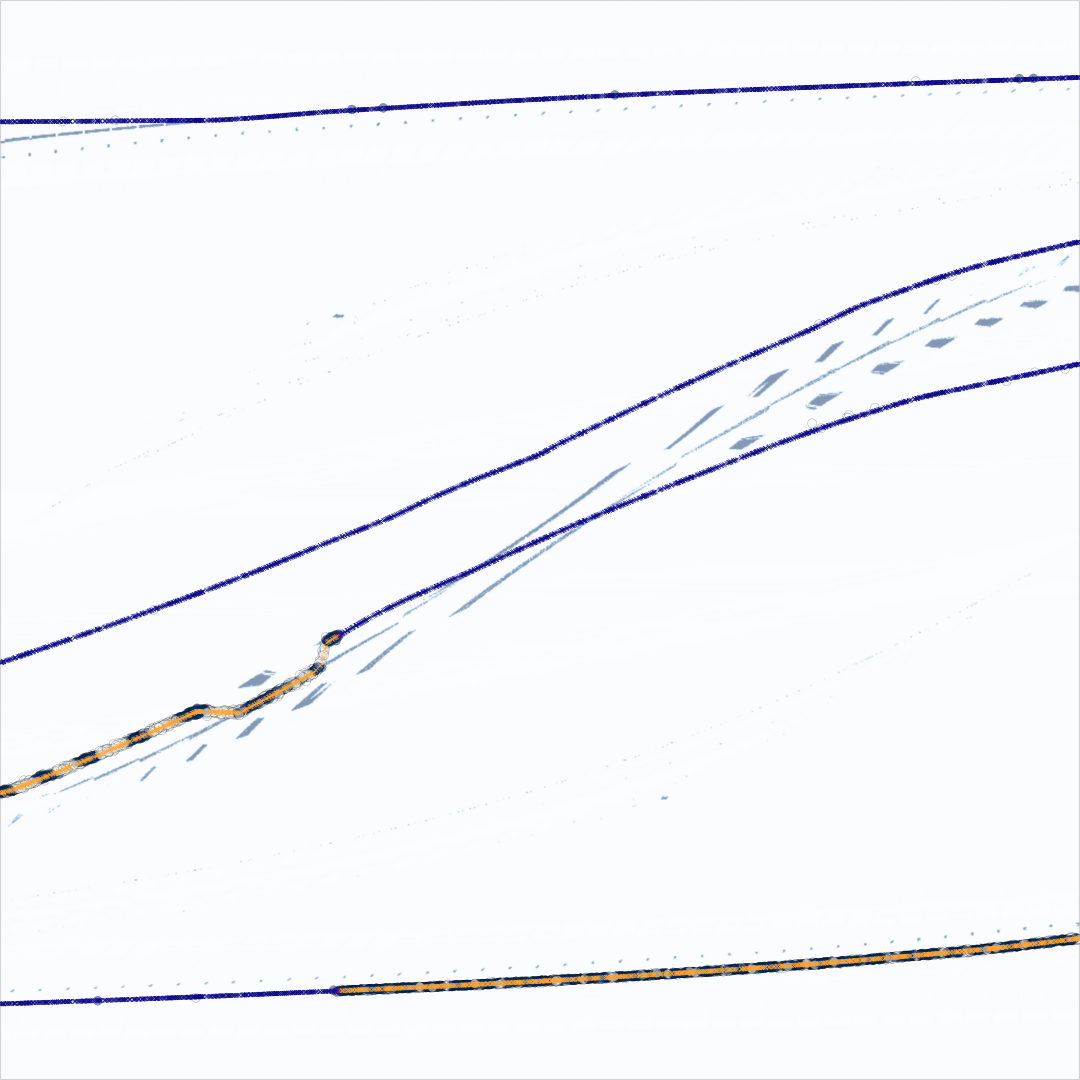}  
    \caption{MoE\\+~distance reg.}\label{fig:moedistancereg}
    \end{subfigure}\hfill
    \begin{subfigure}[t]{0.18\textwidth}
    \centering
    \captionsetup{justification=centering}
    \frame{
    \includegraphics[width=1.0\textwidth]{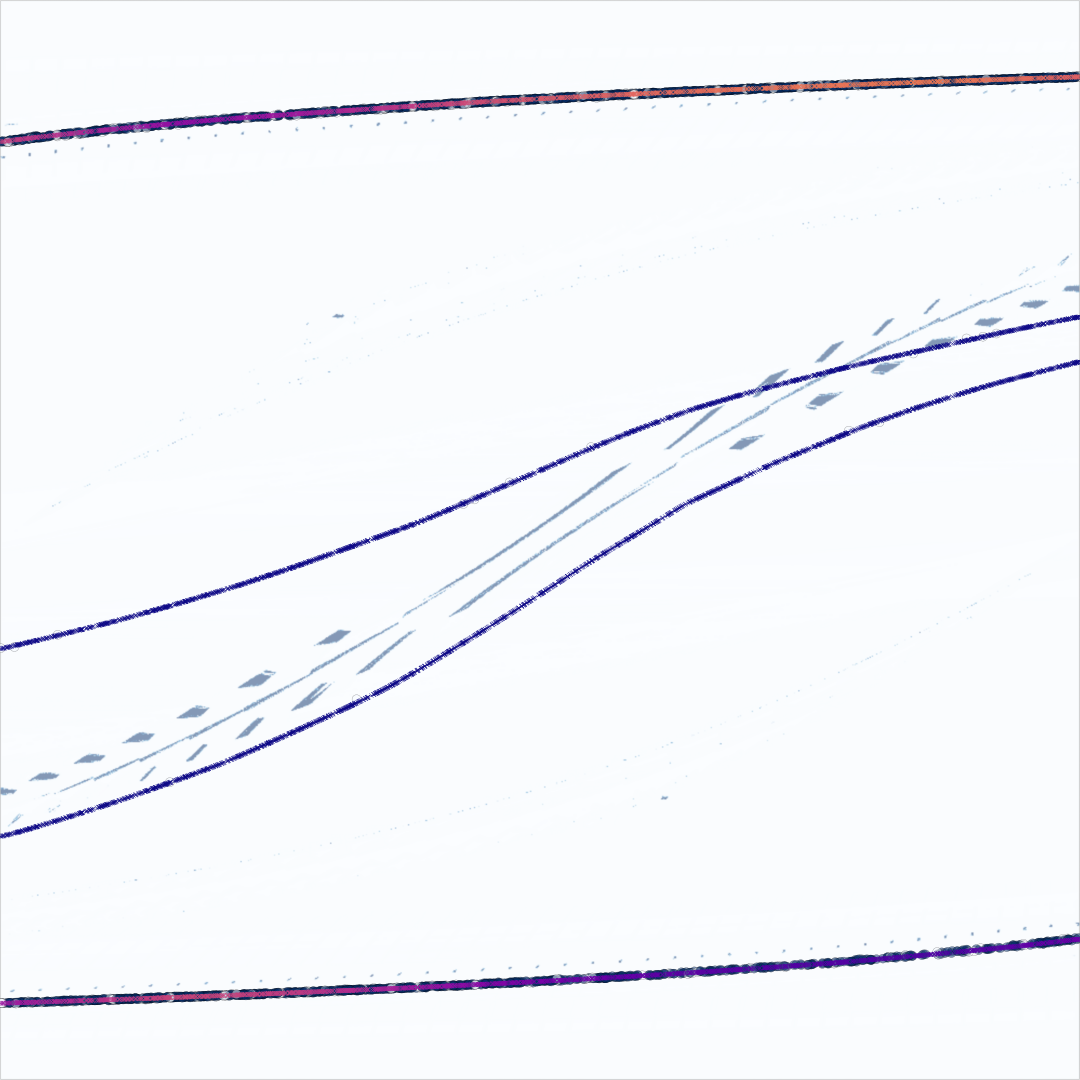}  
    }
    \caption{SCOOT\\(elite + MoE dist. reg. + curriculum)}\label{fig:fullscoot}
    \end{subfigure}\hfill
    \includegraphics[height=0.18\textwidth]{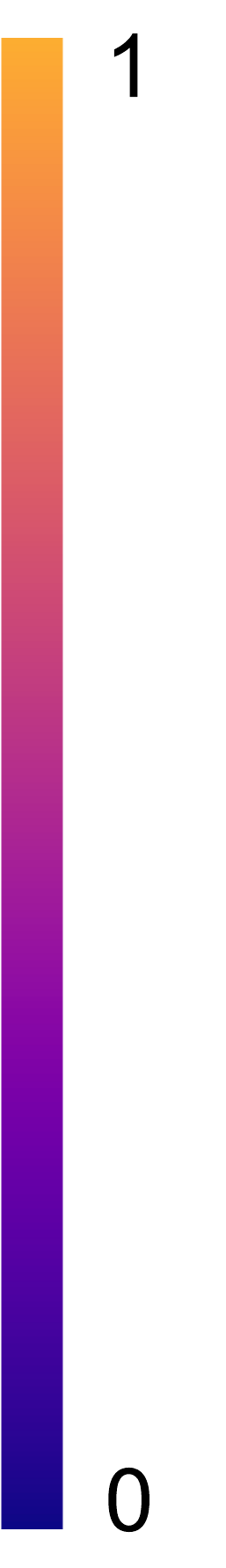}
    \hfill
    \caption{State-conditioned action means across the reward landscape, for each algorithm version. The colors of the means indicate learned mixture component weights $p_h(\bs)$, ranging from 0 (purple) to 1 (orange).}
    \label{fig:incremental}
\end{figure*}

\begin{figure}
    \centering
    \includegraphics[width=0.99\columnwidth]{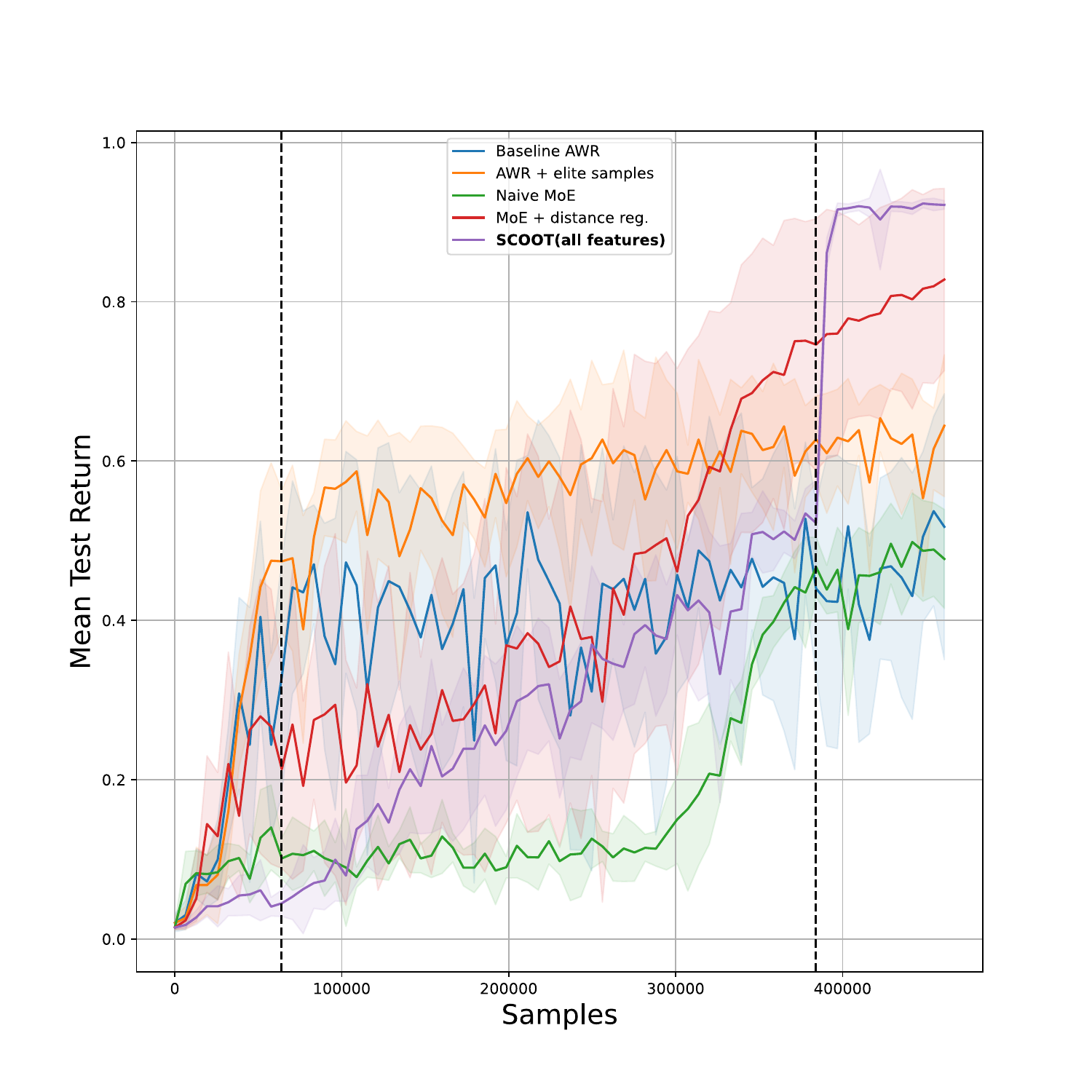}
    \caption{Evaluating algorithm versions in terms of mean test return (approximately equal to shot success rate). The curves show means and standard deviations across 10 random seeds and 2048 randomly sampled states per seed. Vertical dashed lines indicate the stages of our training curriculum (\Cref{sec:curriculum}).}
    \label{fig:learning_curve}
\end{figure}

\paragraph{Impact on Performance} Empirically, using the elite samples and directly using advantages as weights improves performance over the baseline AWR, as illustrated in \Cref{fig:learning_curve,fig:incremental}. This also removes the need for AWR's temperature hyperparameter $\beta$.



\subsection{Distance-Regularized Mixture-of-Experts} \label{sec:moe}
To handle the discontinuities and multimodality of the reward landscape, we employ a multimodal policy. Specifically, we use a mixture-of-experts (MoE) policy with $H$ Gaussian heads:
\begin{align}
    \pi_\theta(\ba \ | \ \bs) &= \sum_{h=1}^H p_h(\bs) \mathcal{N}\left( \ba ; \mu_\theta^h(\bs), \sigma^2 \right), \label{eq:moe_prob}
\end{align}
where $\mathcal{N}$ denotes a Gaussian PDF and $p_h(\bs) \in [0,1]$ is the state-conditioned component weight, i.e., the categorical probability of selecting head $h$ when sampling an action for state $\bs$, with $\sum_{h=1}^H p_h(\bs) = 1$. Each head $h$ corresponds to an individual state-conditioned Gaussian distribution, i.e. a neural network that outputs a state-dependent mean $\mu_\theta^h \left( \bs \right)$. We employ a global scalar variance $\sigma^2$ (shared across heads and states) which is learned together with the neural network parameters. 

\paragraph{Distance Regularization} As illustrated in \Cref{fig:naivemoe}, the MoE policy is not optimal on its own, as the head means may all converge to the same central reward modes, failing to explore the action space fully. To encourage the policy heads to avoid overlap and cover diverse parts of the state-action space, we add the following Mahalanobis distance regularization term to the policy optimization objective (\Cref{alg:scoot} Line \ref{algline:policy}): 
\begin{align}
    \mathcal{L}_{\text{dist}}\left( \theta, \bs \right) &= \max\left(0, 1 - \frac{1}{d}\min_{h_1 \neq h_2} \frac{||\mu_\theta^{h_1}(\bs)-\mu_\theta^{h_2}(\bs)||}{\sigma}\right) \label{eq:dist_loss}
\end{align}
where $h_1$ and $h_2$ enumerate the indices of Gaussian heads of the MoE policy, and $d$ (a hyperparameter) is the minimum amount of overlap to accrue penalty, i.e. the $x$-intercept of the decreasing linear function clipped at 0. 

\paragraph{Non-overlapping Initialization} The distance regularization may be ineffective or unstable if the policy heads are initialized with high overlap. Therefore, we initialize each policy head such that, for all states, the action means are distinct and cover the whole action space. Denoting head $h$'s action mean as $\mu_h$, we adjust the policy network's final layer output $\bz_h$ as:
\begin{align}
\mu_h=\alpha_h \bz_h + \beta_h \label{eq:head_init}
\end{align}
where $\alpha_h \in \mathbb{R}$ is a scale parameter, and $\mathbf{\beta}_h \in \mathbb{R}^{|\ba|}$ is a shift parameter corresponding to head $h$. Both $\alpha_h$ and $\beta_h$ are adapted through the policy optimization. The shift parameters $\mathbf{\beta}_1, \cdots, \mathbf{\beta}_H$ are initialized by a $|\ba|$-dimensional space-filling Sobol sequence \cite{sobol1967distribution} of length $H$, and $\alpha_h$ is initialized to a low value (0.001) so that initially, $\mu_h \approx \mathbf{\beta}_h$ independent of $\bz_h$.


\paragraph{Impact on Performance} As illustrated in \Cref{fig:learning_curve,fig:moedistancereg}, adding the distance regularization with the non-overlapping initialization improves results. The policy heads now cover the whole action space, and the final policy return is higher, although the single Gaussian policy (AWR + elite samples) learns faster initially. 



\begin{figure*}
    \centering
    \begin{subfigure}[t]{0.31\textwidth}
    \centering
    \captionsetup{justification=centering}
    \includegraphics[width=\textwidth]{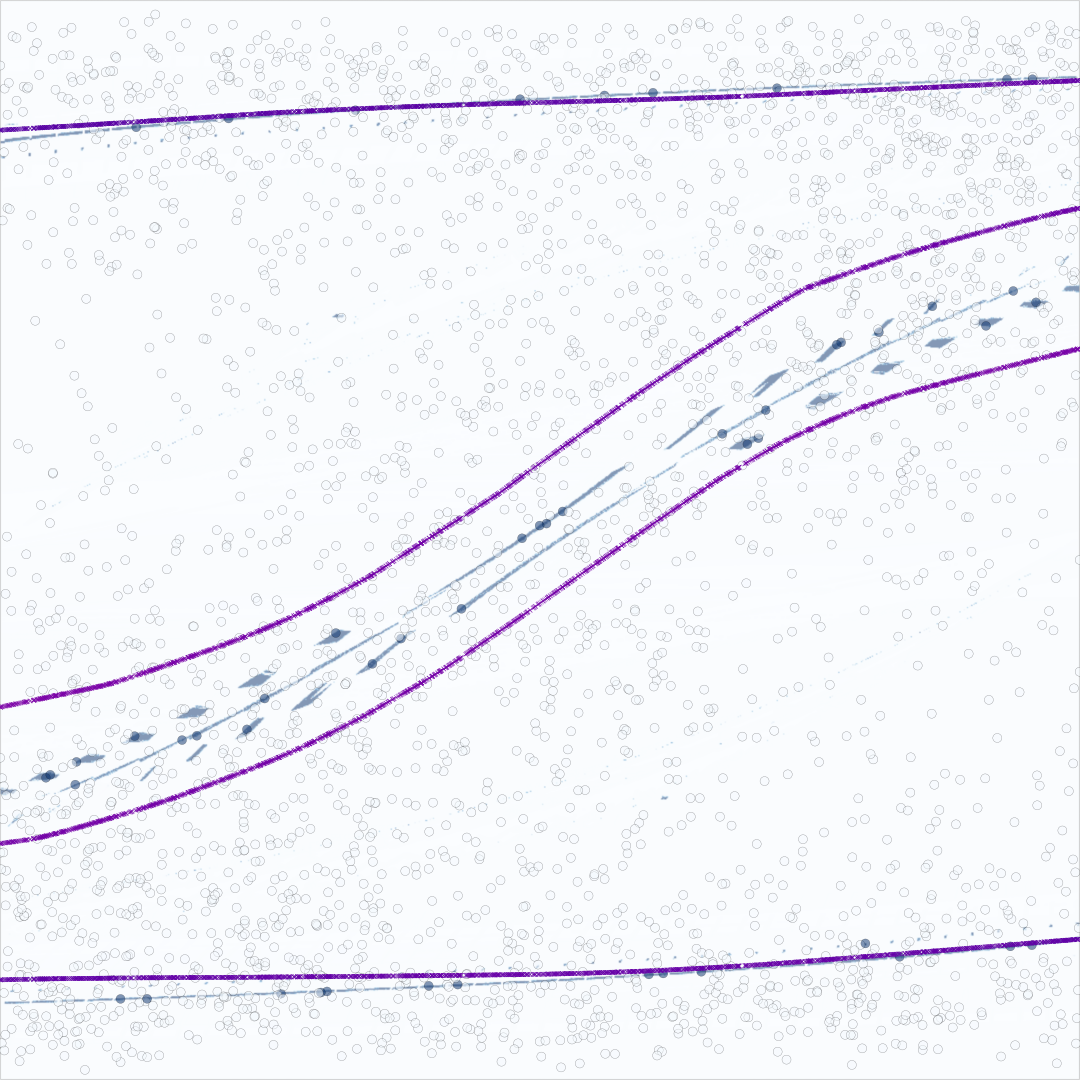}
    \caption{Before learning action variance}
    \label{fig:curriculum_1}
    \end{subfigure}\hfill
    \begin{subfigure}[t]{0.31\textwidth}
    \centering
    \captionsetup{justification=centering}
    \includegraphics[width=\textwidth]{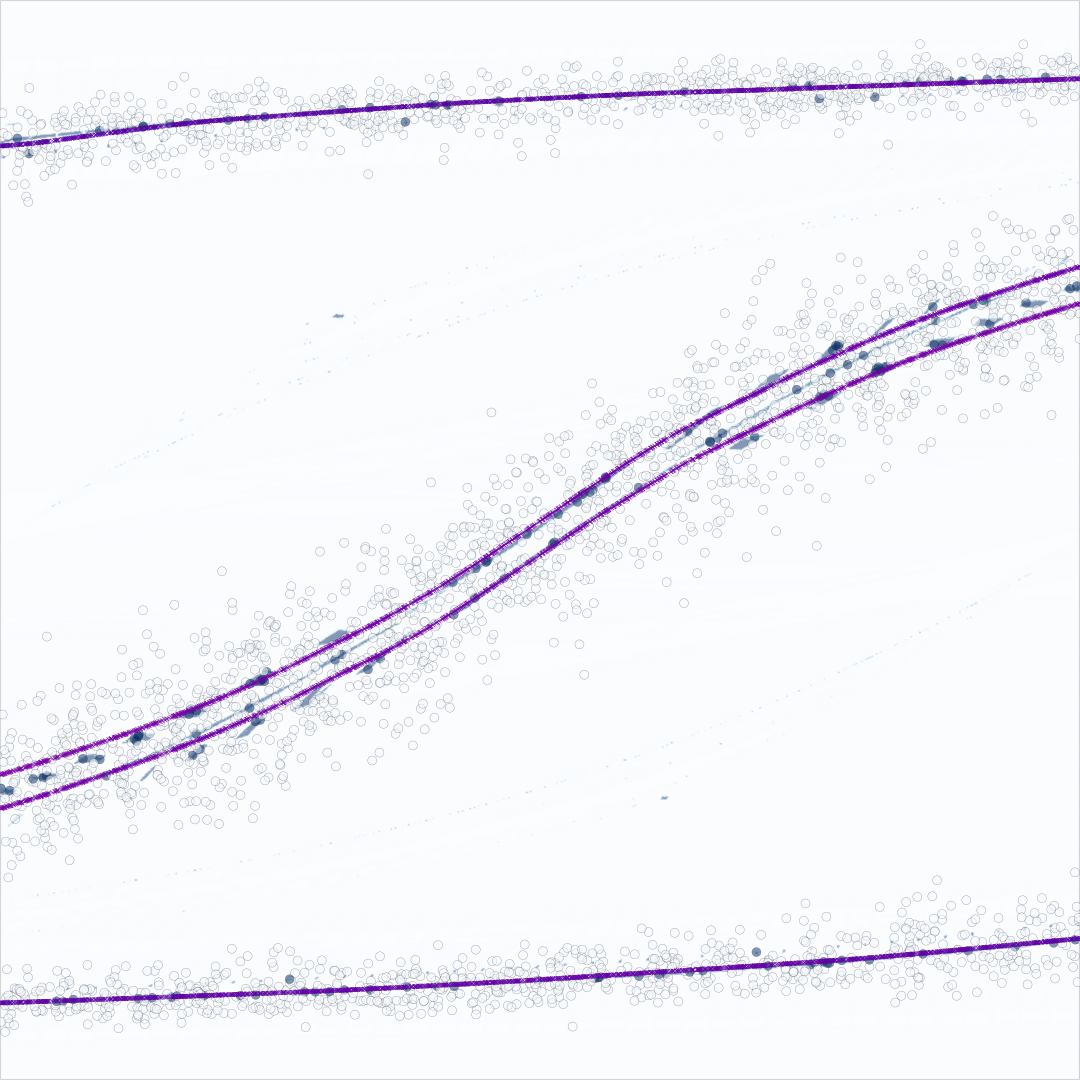}
    \caption{Before learning component weights}
    \label{fig:curriculum_2}
    \end{subfigure}\hfill
    \begin{subfigure}[t]{0.31\textwidth}
    \centering
    \captionsetup{justification=centering}
    \includegraphics[width=\textwidth]{figures/scoot_converged.png}
    \caption{After learning component weights}
    \label{fig:curriculum_3}
    \end{subfigure}\hfill
    \includegraphics[height=0.31\textwidth]{figures/colorbar.pdf}
    \hfill
    \caption{SCOOT's action samples and means throughout the learning curriculum. The colors of the means indicate learned mixture component weights $p_h(\bs)$, ranging from 0 (purple) to 1 (orange).}
    \label{fig:curriculum}
\end{figure*}


\subsection{Training Curriculum} \label{sec:curriculum}

With the features above, our initial experiments indicated two common failure modes:

\begin{itemize}[leftmargin=*]
    \item Before the experience replay buffer is filled, initial convergence often is random and leads to unrecoverable errors.
    \item Most of the MoE component weights may rapidly decay to zero, leaving much of the state-action space underexplored. This can cause results like \Cref{fig:moedistancereg}, where the dominant head attempts to model the highly discontinuous central reward ridge, leading to poor actions in many states.
\end{itemize}

To prevent such failure modes, we implement curriculum learning \cite{bengio2009curriculum} with the following three stages:
\begin{enumerate}[leftmargin=*]
    \item In the first stage (\Cref{fig:curriculum_1}), only the action means are learned while the action variance and the component weights are held constant. This ensures that the experience replay buffer has sufficient information for learning before substantial adaptation occurs, while also primitively adapting towards local optima that are discoverable early on.
    \item In the second stage (\Cref{fig:curriculum_2}), the action variance is also allowed to reduce from its initially high value, allowing the policy heads to model the reward optima more accurately. 
    \item Finally, in the third stage (\Cref{fig:curriculum_3}), the state-conditioned component weights are also learned. This effectively ``shuts down'' policy heads that are not sufficiently close to any local optimum, while redirecting the sampling budget to the policy heads that are near local optima. The component weights are hence learned to choose the best actions found for each state.
\end{enumerate}

\paragraph{Impact on Performance} As illustrated in \Cref{fig:learning_curve,fig:fullscoot,fig:curriculum}, adding the curriculum improves the mean return and allows the policy to correctly model the continuous ridges at the top and bottom of the landscape. The policy correctly models most of the high-reward regions at the end of the second stage (\Cref{fig:curriculum_2}), although the two central heads start to drift randomly when their weights decay in the third stage (\Cref{fig:curriculum_3}). The loss of modalities appearing in \Cref{fig:curriculum_3}, however, does not affect the performance of the final policy, as it chooses its actions only from the active components. One may prevent this modality loss by further regularizing the learned component weights by, e.g., blending the $p_h$ values with uniform probability.


\subsection{Implementation Details}\label{sec:implementation}

We use PyTorch \cite{paszke2019pytorch} to implement the neural networks. The policy network is a multi-layered perceptron (MLP) of hidden dimensions $128 \times 64$ with (\texttt{ReLU}) activations. This is identical to the architecture used in AWR \cite{peng2019advantage}. The final actions are bounded in $[-1, 1]$ with $\tanh$ transformations as implemented in SAC \cite{haarnoja2018soft} (which AWR does not use). The initial action variance is set to $\log\sigma = -1$.
The value network uses the same architecture minus the $\tanh$ transformations.
The state-conditioned categorical weights $p_1(\bs), \cdots, p_h(\bs)$ are also computed by a similar MLP, whose final output is transformed into a categorical distribution via softmax. 
We use RAdam \cite{liu2019variance} optimizer to train our networks. The elite sample selection allows for a relatively aggressive learning rate of 1e-3 for the policy. For the value network, the learning rate is 1e-5. Given that simulating each action (a billiards shot with 200 physics timesteps) is slow, we restrict our iteration budget to 128 samples per iteration. The experience replay buffer is a FIFO queue with a maximum size of 3200 samples, i.e., we use data from the past 25 iterations for updating the networks, which is also AWR's default setting. We train the networks for a single epoch per iteration with mini-batches of 256 samples. Full-batch updates are used when the number of samples in the buffer is less than or equal to 256. The distance penalty weight $\lambda_\text{dist}$ is set to $0.1$ with the overlap limit $d=1.0$ to prevent action means from approaching within one standard deviation from each other. 

As each billiards shot is simulated for 200 physics timesteps, large-scale experiments and evaluations can be excessively slow. Therefore, in the case of 1-dimensional actions and states, training and evaluation are accelerated by using a pre-computed state-action reward look-up table with \verb|~|8.7 million samples. The table is also used to generate the reward landscape figures and the learning curves of \Cref{fig:learning_curve}, while the billiards figures and videos in \Cref{sec:results} use the real simulator. 

Stochastic actions are sampled from the mixture during training to maximize action precision and minimize noise without sacrificing exploration. As multiple Gaussian heads may have similar learned component weights corresponding to multiple reward peaks, the policy generates each action for evaluation as the mean of the Gaussian head sampled according to the $p_h(\bs)$ values.


\section{Results}\label{sec:results}

\textbf{Please refer to the supplementary video section for animated results.} Above, we have already validated the benefits of the proposed SCOOT features, as summarized in \Cref{fig:learning_curve,fig:incremental}. In this section, we showcase the resulting policy's capability in producing high-precision billiards shots and a promising direction in producing diverse shots leveraging the MoE. We also provide an initial outlook toward higher-dimensional generalization of SCOOT. 

\paragraph{Learning High-Precision Shots}

\begin{figure*}
    \centering
    \begin{subfigure}[t]{0.49\textwidth}
    \centering
    \includegraphics[width=0.3\textwidth,trim={940 120 940 120},clip]{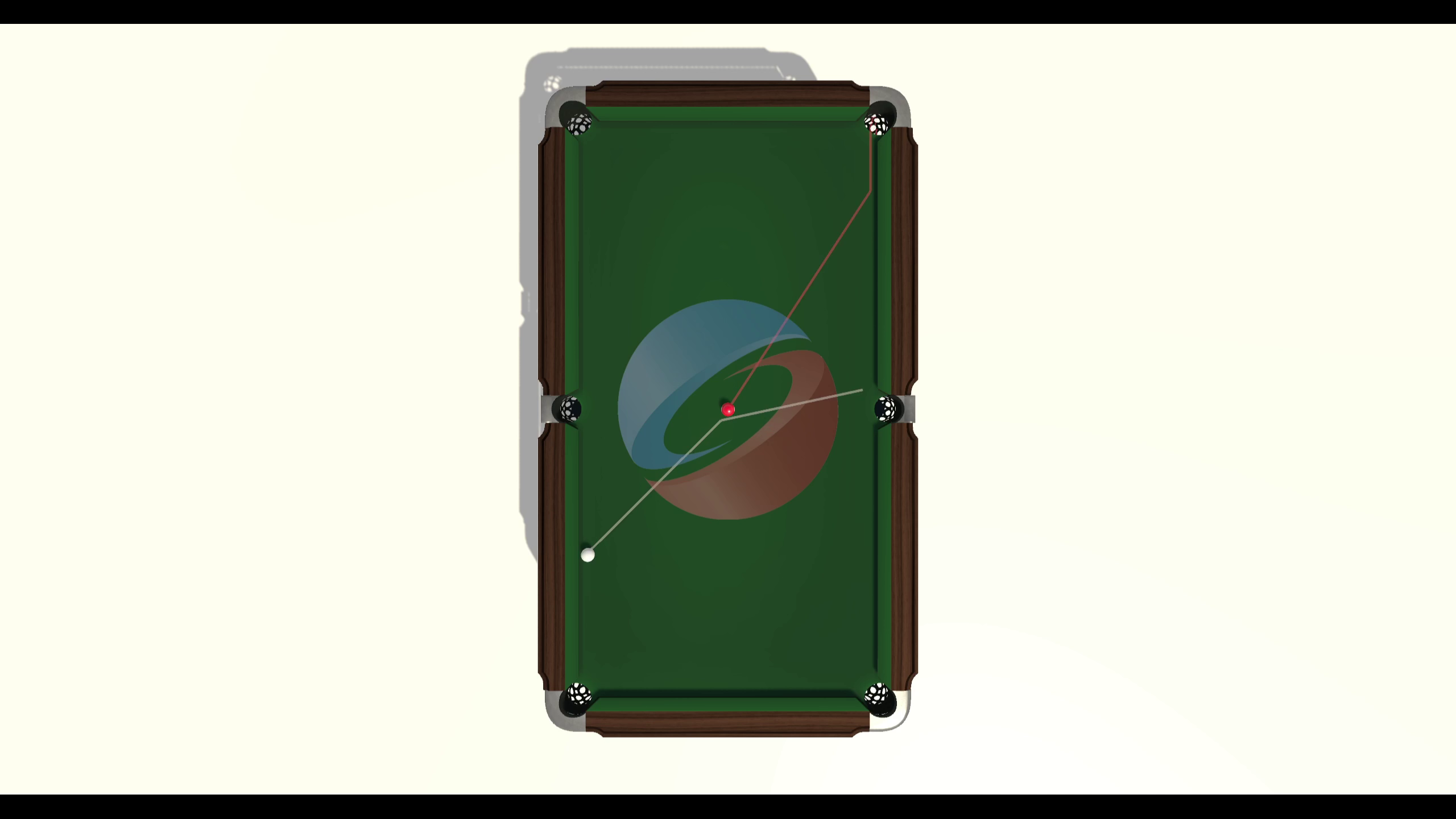}
    \includegraphics[width=0.3\textwidth,trim={940 120 940 120},clip]{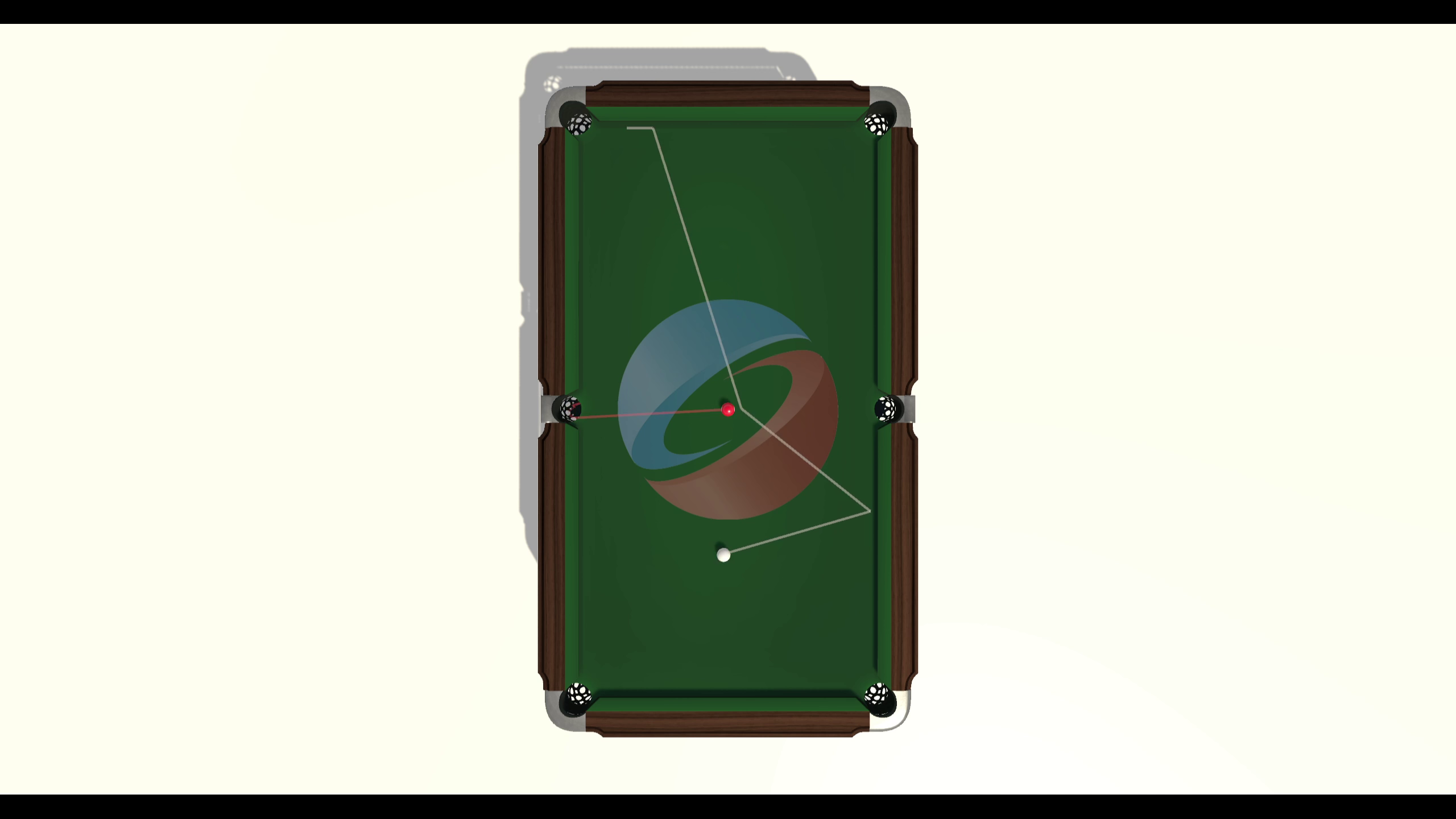}
    \includegraphics[width=0.3\textwidth,trim={940 120 940 120},clip]{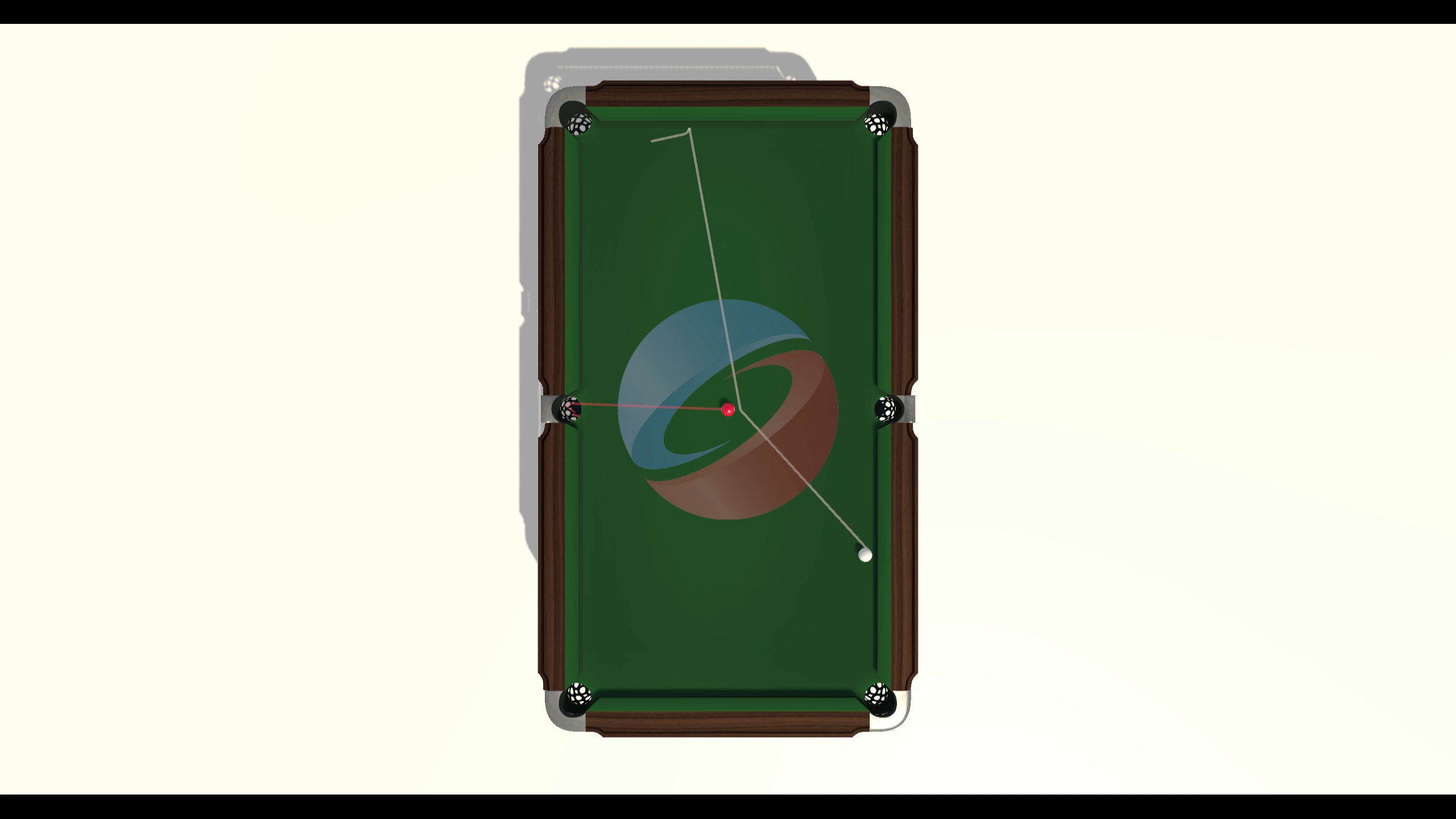}
    \caption{Visualization of high-precision shots.}
    \label{fig:precise}
    \end{subfigure}
    \begin{subfigure}[t]{0.49\textwidth}
    \centering
    \includegraphics[width=0.3\textwidth,trim={700 100 700 100},clip]{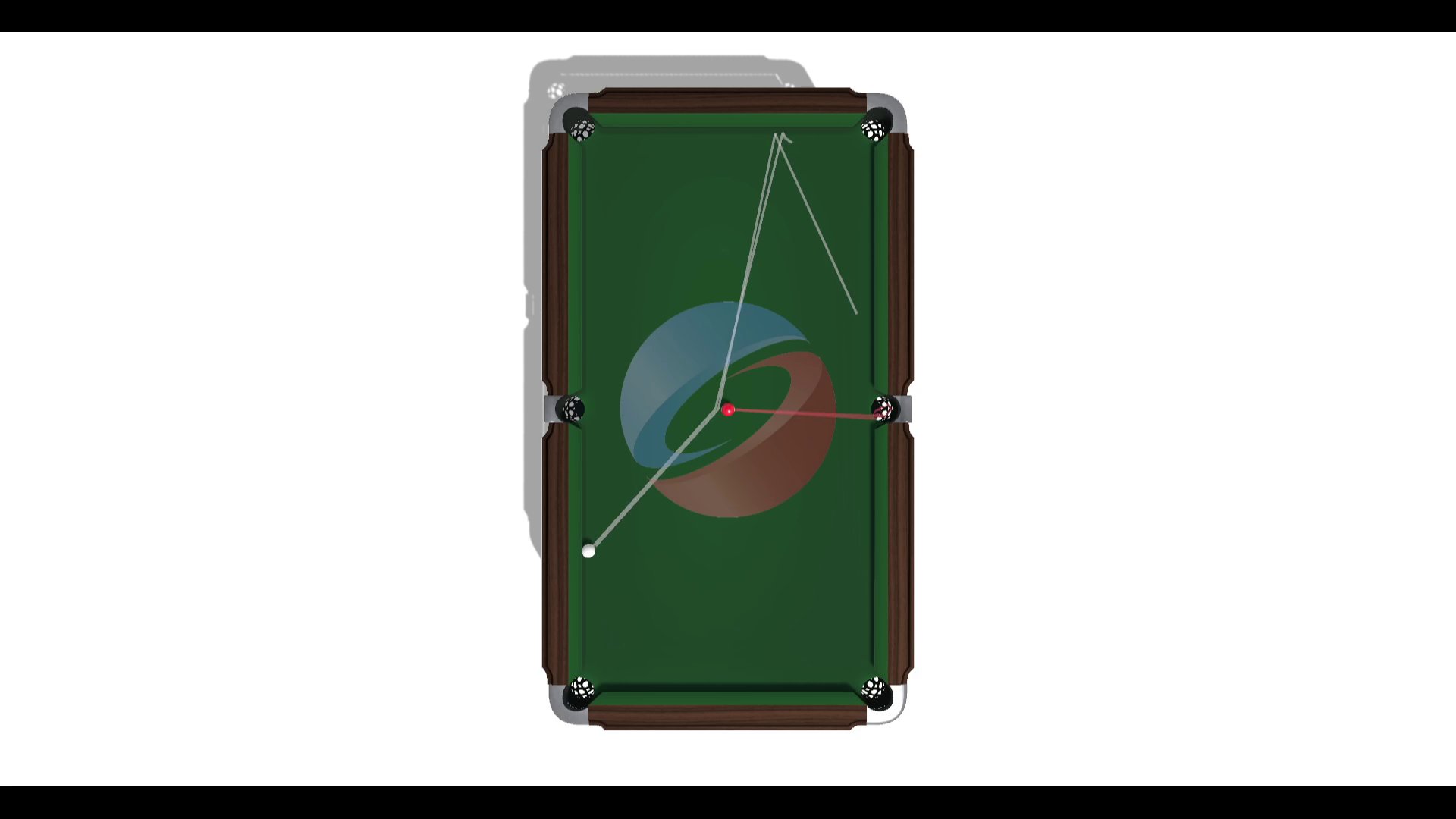}
    \includegraphics[width=0.3\textwidth,trim={700 100 700 100},clip]{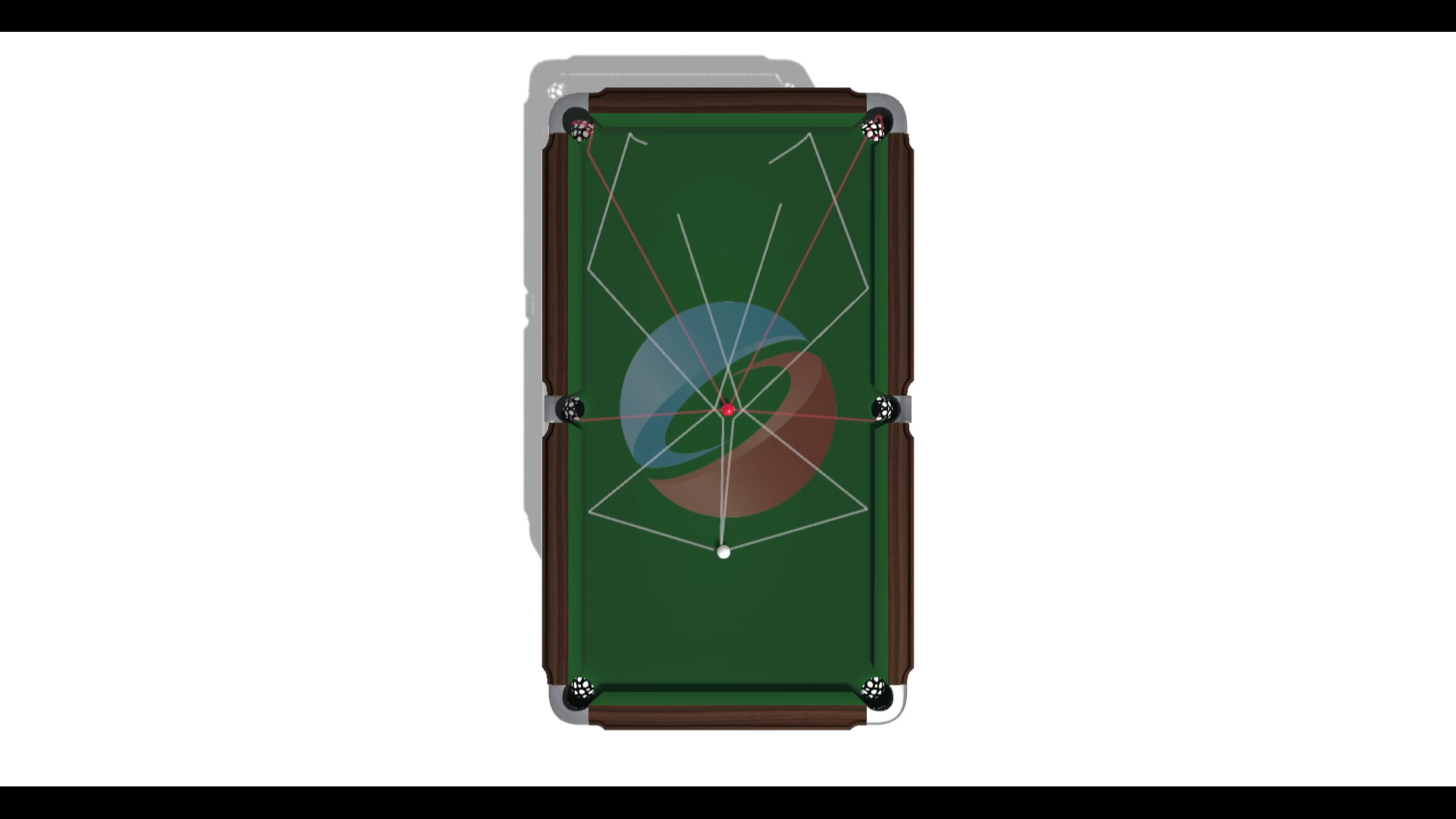}
    \includegraphics[width=0.3\textwidth,trim={700 100 700 100},clip]{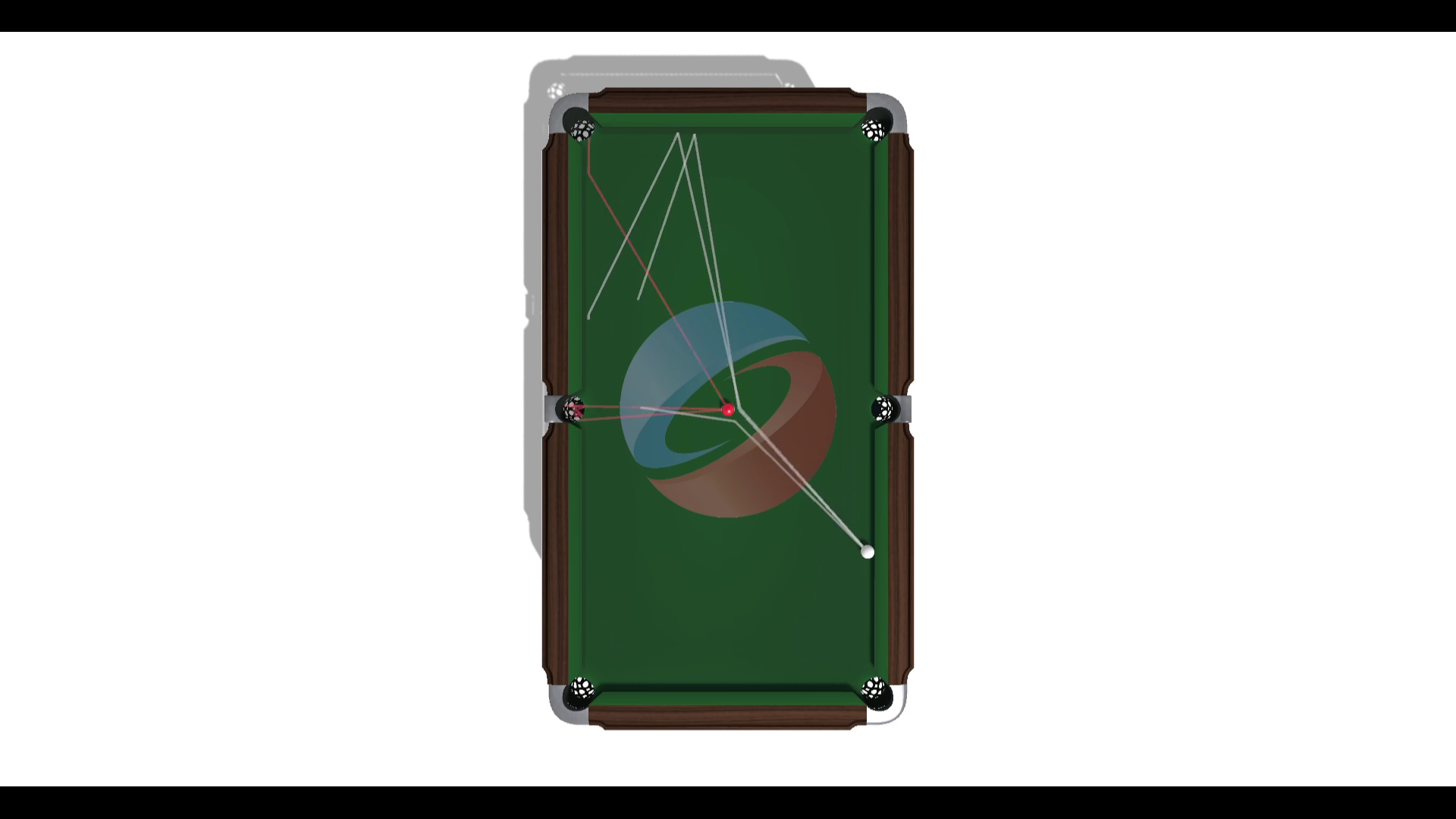}
    \caption{Discovery of diverse solution modes.}
    \label{fig:diverse}
    \end{subfigure}
    \caption{Successful actions with (a) precision and (b) diversity are discovered across various states. The white lines show the cue ball's trajectory, and the red lines the target ball's.}
\end{figure*}

\Cref{fig:precise} showcases the capability of a policy learned using SCOOT: we successfully discover a control policy for 1-dimensional billiards shots, switching between strategies as necessary. Given the limited capacity of the neural network policy, we observe a keen bias towards continuous and more rigid reward peaks. For example, \Cref{fig:curriculum_3} shows that the central reward peaks, initially modeled during the intermediate stages in the curriculum, are no longer used after component weights are learned. This corroborates our intuition that the agent's actions default to risk-averse solution modes in the presence of high-precision requirements. With sufficient rigidity and continuity, the agent can exploit known solutions from other states to a large extent and thus produce a high-return action for the current state. This also highlights the importance of sufficient exploration; we discover such solution modes via SCOOT's MoE policy, distance regularization, and curriculum learning features. Policies trained without these features tend to fail to discover these rigid and continuous reward peaks consistently (\Cref{fig:incremental}). The discovery of such reward peaks explains the proposed features' superiority in sample efficiency, observed test returns, and sensitivity to initialization (\Cref{fig:learning_curve}).

\paragraph{Discovering Diverse Shots}
In our test problem, SCOOT can discover various local reward peaks across the landscape and converge the action means towards them. For example, \Cref{fig:curriculum_2} shows that the 4-headed policy closely follows the shape of the 4 major reward peaks of the billiards environment. Continuing to train the policy from this state without optimizing component weights, we observe that the policy converges towards these local peaks. Visualizing the converged action means of the policy, we observe that this corresponds to producing diverse shots, with each shot capturing a unique strategy while scoring a high return. \Cref{fig:diverse} showcases the capabilities of such a policy, which outputs a variety of shots for each input state. 

In our specific problem setup, there are 4 distinct strategies in almost every state, which can be modeled with a MoE policy with 4 heads. In the general case, future work may explore techniques for learning the number of viable strategies, e.g., by using a per-head value network that could be queried during inference to predict the return for the strategy corresponding to each head.


\paragraph{Scaling Up}


\begin{figure*}
    \centering
    \begin{subfigure}[t]{0.49\textwidth}
    \centering
    \includegraphics[width=0.3\textwidth,trim={1000 75 1000 75},clip]{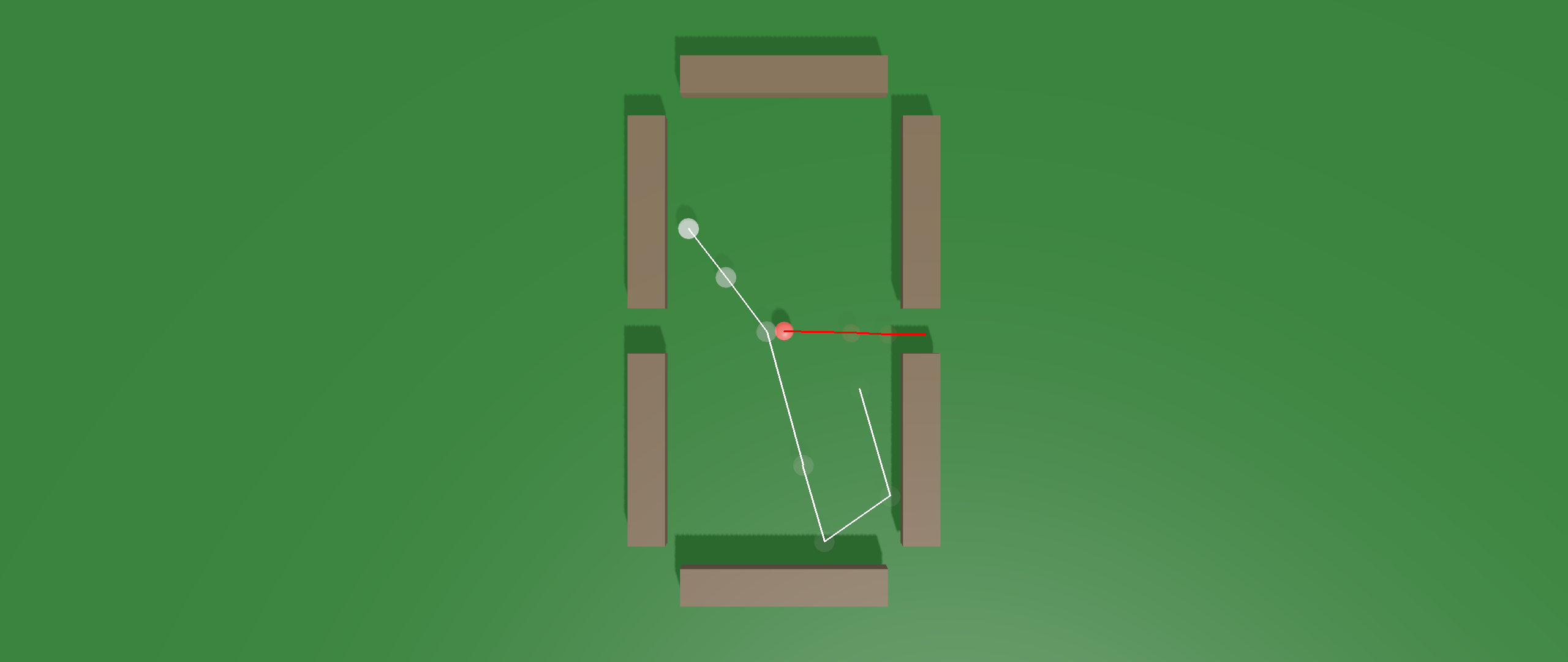}
    \includegraphics[width=0.3\textwidth,trim={1000 75 1000 75},clip]{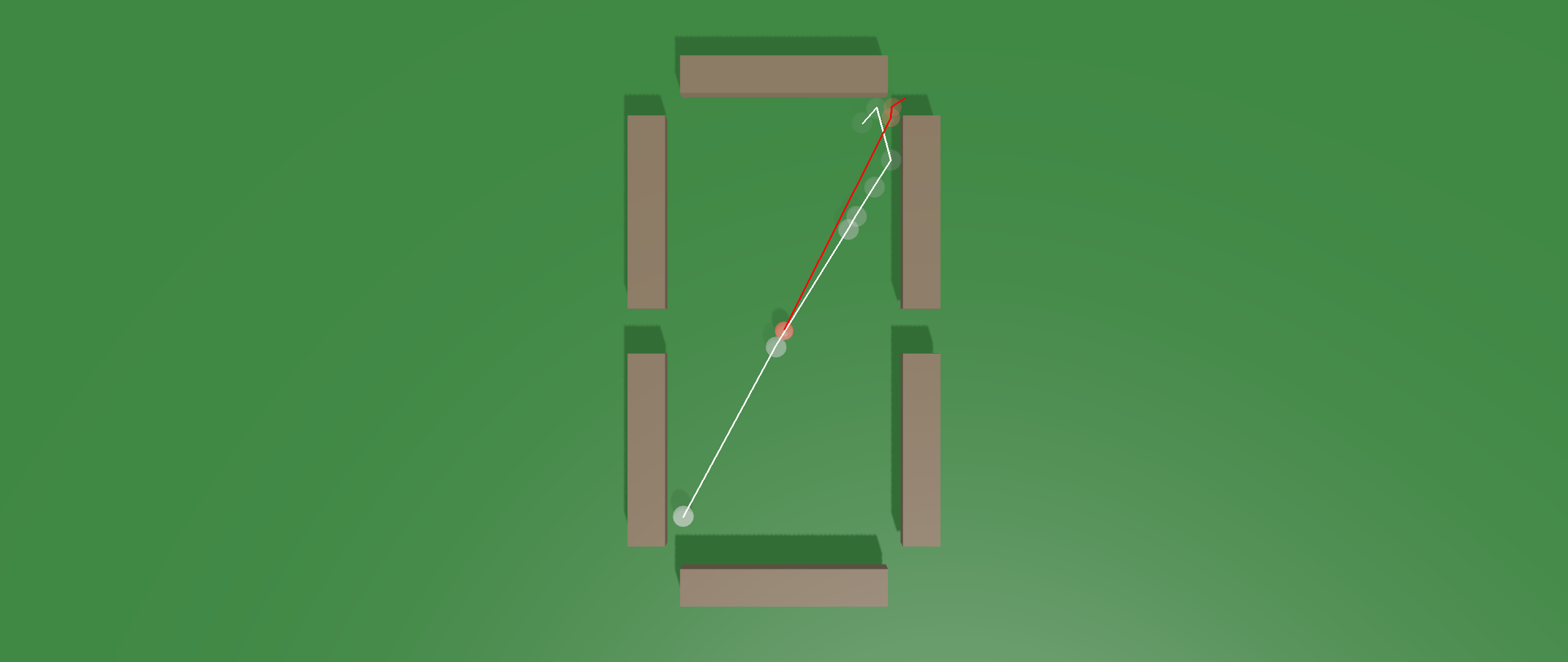}
    \includegraphics[width=0.3\textwidth,trim={1000 75 1000 75},clip]{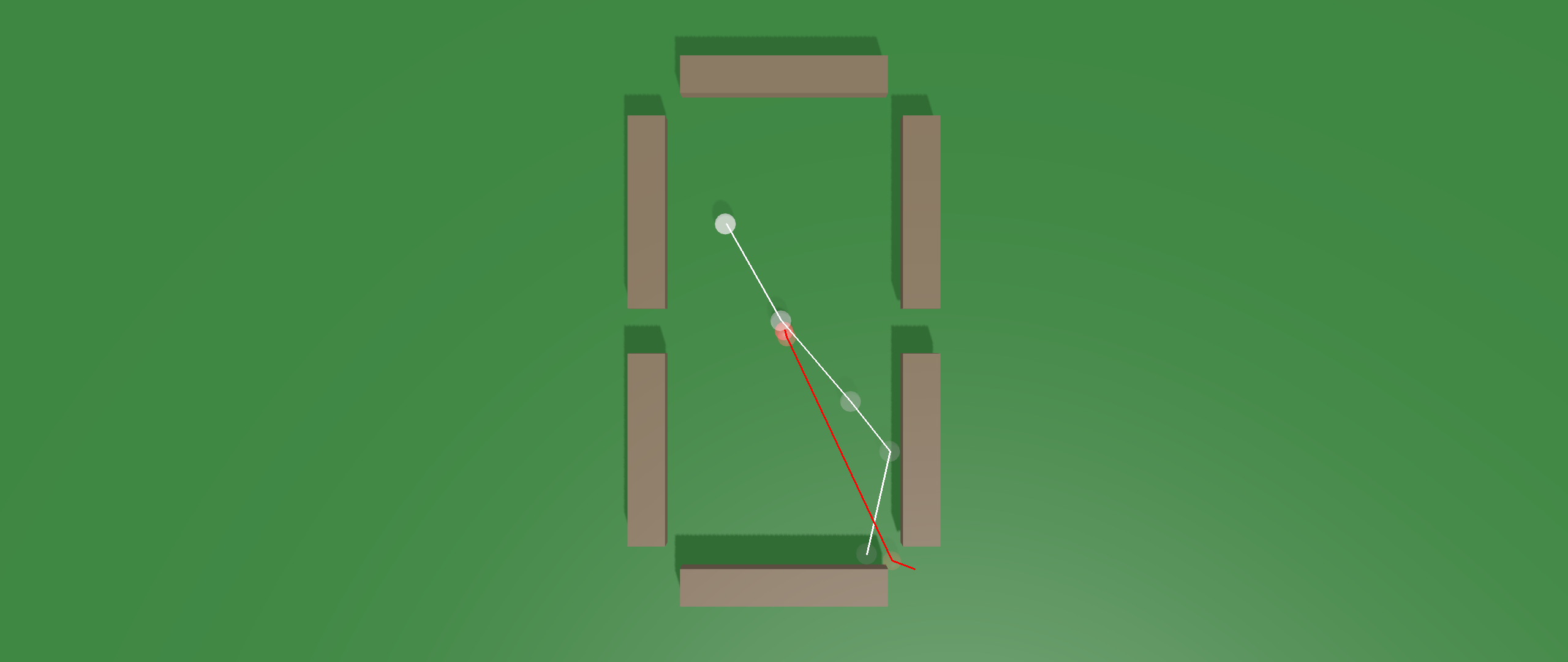}
    \caption{Successful shots.}
    \label{fig:2s2a_good}
    \end{subfigure}
    \begin{subfigure}[t]{0.49\textwidth}
    \centering
    \includegraphics[width=0.3\textwidth,trim={1000 75 1000 75},clip]{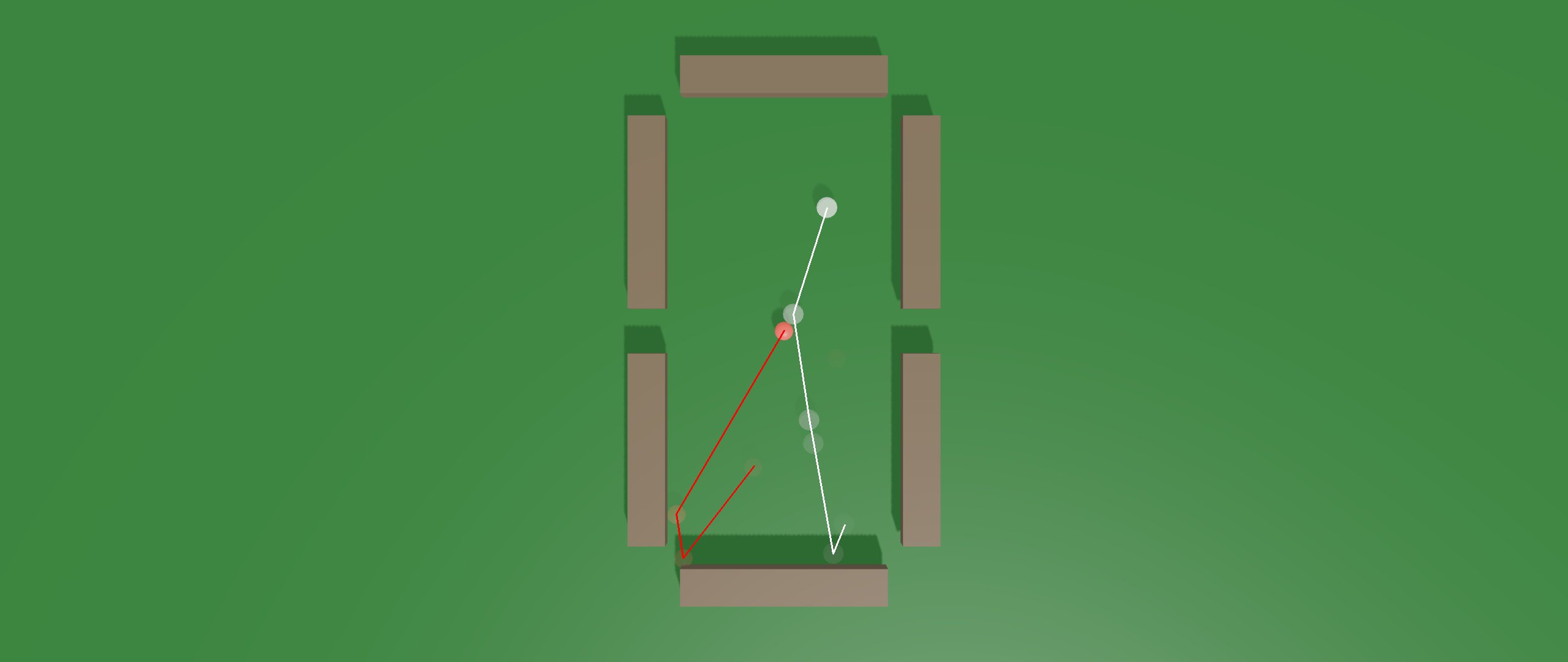}
    \includegraphics[width=0.3\textwidth,trim={1000 75 1000 75},clip]{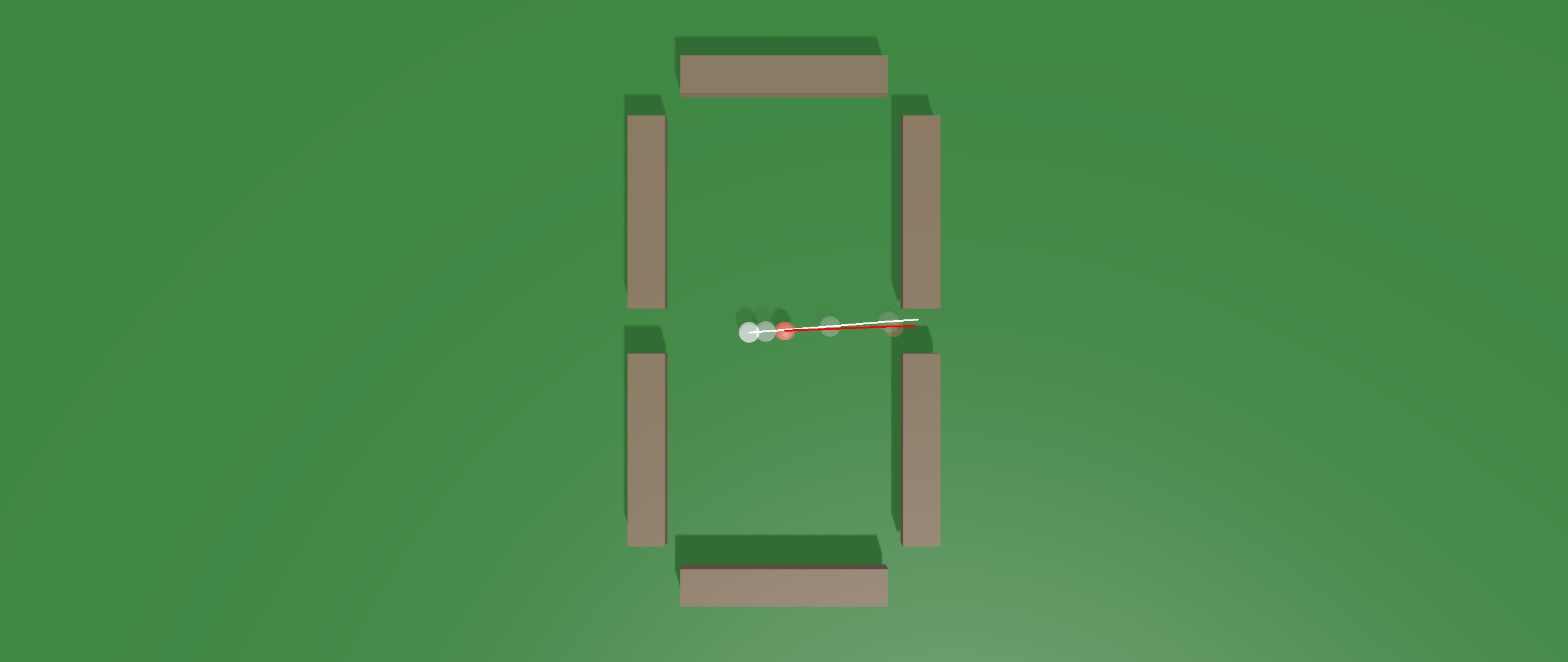}
    \includegraphics[width=0.3\textwidth,trim={1000 75 1000 75},clip]{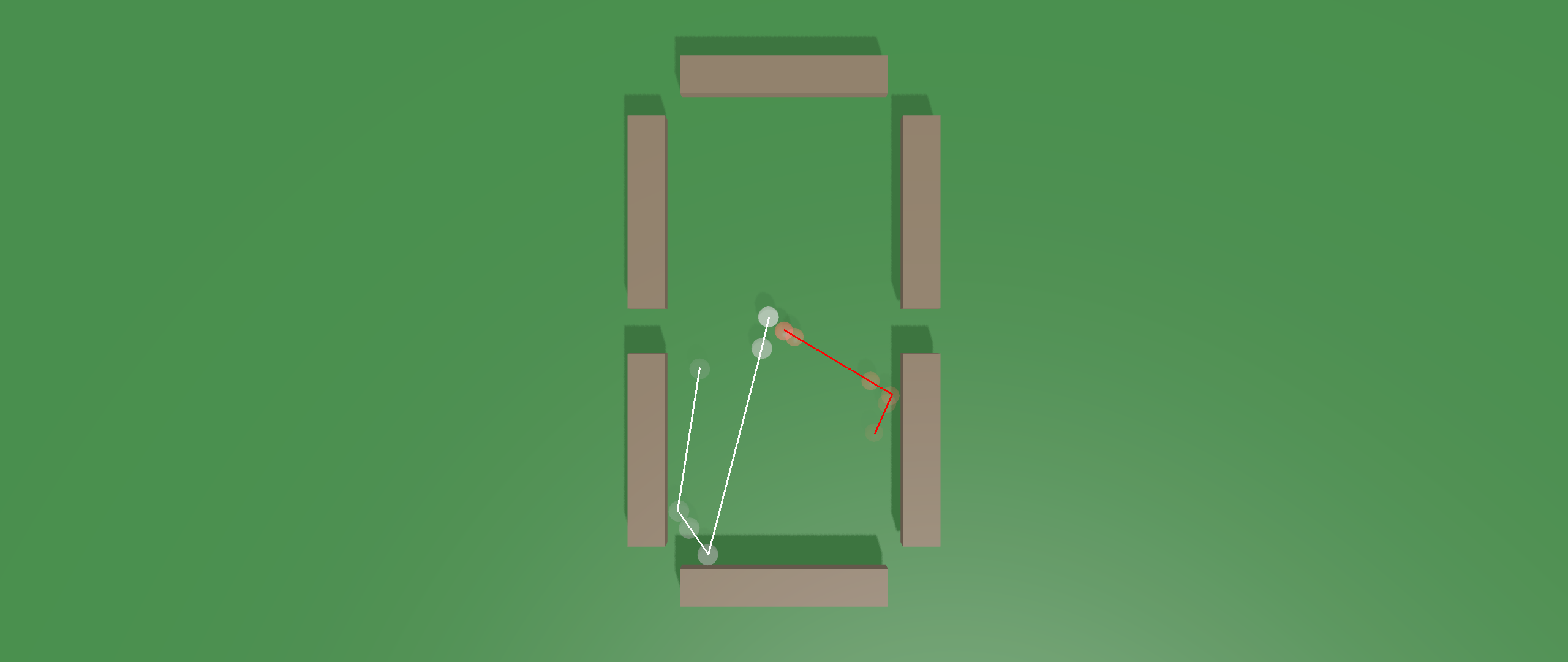}
    \caption{Failed shots.}
    \label{fig:2s2a_bad}
    \end{subfigure}
    \caption{(a) Successful and (b) failed shots by SCOOT in billiards with 2-dimensional state and action parameterization.}
\end{figure*}


To assess the generalization of our method, we use SCOOT on a slightly higher-dimensional problem, where the cue ball's position varies in both $x$- and $y$-directions. The action is parameterized as $\ba = (f_x, f_y)$ (a 2-dimensional force vector).
Increasing dimensionality makes it exponentially more difficult to fill out the state-action space, making the lookup table acceleration (\Cref{sec:implementation}) infeasible due to the amount of memory required. To this end, we instead implement a GPU-accelerated billiards simulator using Brax \cite{brax2021github}. Simpler geometries are used instead of a true mesh model of the pool table to reduce simulation overhead. For training the agent for this environment, we set the number of Gaussian heads $H = 8$ to encourage the discovery of more local peaks in the higher-dimensional state-action space.

Without further modifications, training for 72 hours on an NVIDIA A100 Tensor Core GPU, SCOOT discovers a policy that can perform reasonably well (\Cref{fig:2s2a_good}, success rate around 75\%), although occasionally failing due to imprecise actions (\Cref{fig:2s2a_bad}).

It is evident that SCOOT iterations discover local reward peaks and that the policy adapts to them, albeit not completely. For example, most failure cases in \Cref{fig:2s2a_bad} are due to excessively strong force or a slight misalignment in shot angles. Additional training with the RAdam optimizer seems to increase performance gradually, but we observe that adaptation slows down to an infeasible amount. This challenge may necessitate further modifications on top of the ones proposed in this work, e.g., learning rate adaptation or a PPO-style KL-divergence limit.

It is easy to increase the dimensionality of the problem further by introducing multiple target balls and allowing their positions to vary. So far, we have not yet achieved a solidly performing policy in such problem variants. More experiments are needed, and we conjecture that a compounded effect of the curse of dimensionality, limited policy capacity, and the even more irregular reward landscape due to additional collision possibilities may be at play. A competent generalization to such high-dimensional variations of billiards will remain an open challenge for researchers investigating high-risk, high-precision motion control. However, we believe our work can provide some useful building blocks.

\section{Conclusion}


We have proposed SCOOT, a novel DRL algorithm for learning to perform high-precision and high-risk movement control tasks. As SCOOT inherits the simplicity of AWR with only minor modifications, we hope our work will find use in various applications that require highly precise movement control. Additionally, we have demonstrated the capability of SCOOT to learn multiple alternative action strategies. Inspired by this, we are currently working towards applying and extending SCOOT to physically based character animation tasks combining high skill with creative and surprising movements.

We have evaluated SCOOT in the highly challenging problem of learning billiard shots, albeit limited to relatively simple problem variants. However, even these variants turn out to be difficult for commonly used DRL algorithms, demonstrating the merits of SCOOT in unlocking potential avenues for investigating the under-explored class of high-risk and high-precision motion control problems.

\begin{acks}
We would like to thank Christian Guckelsberger for providing invaluable discussions throughout the early stages of this work.
The anonymous MIG reviewers have provided excellent feedback, which we have incorporated into the final revision.
This project is partly funded by the Finnish Center for Artificial Intelligence (FCAI) and the Natural Sciences and Engineering Research Council (NSERC) of Canada.
\end{acks}

\bibliographystyle{ACM-Reference-Format}
\bibliography{references}

\end{document}